\documentclass[a4paper]{cas-sc}

\usepackage[numbers]{natbib}
\usepackage{bm}
\usepackage{pifont}
\usepackage{float}
\usepackage{subfigure}
\usepackage{amsmath,amssymb}
\usepackage{graphicx}
\usepackage{caption}
\usepackage{booktabs}
\DeclareMathOperator*{\argmin}{arg\,min} 
\usepackage{threeparttable}
\usepackage{hyperref}

\def\tsc#1{\csdef{#1}{\textsc{\lowercase{#1}}\xspace}}
\tsc{WGM}
\tsc{QE}
\begin{document}
\let\WriteBookmarks\relax
\def\floatpagepagefraction{1}
\def\textpagefraction{.001}
\let\printorcid\relax % 可去掉页面下方的ORCID(s)

% \shorttitle{<short title of the paper for running head>}    
% \shorttitle{MLF-4DRCNet: Multi-Level Fusion with 4D Radar and Camera for 3D Object Detection in Autonomous Driving}   

% \shortauthors{<short author list for running head>} 
% \shortauthors{Y. Wu \textit{et~al.}}

\shorttitle{}

\shortauthors{Y. Wu \textit{et~al.}}

\title[mode = title]{MLF-4DRCNet: Multi-Level Fusion with 4D Radar and Camera for 3D Object Detection in Autonomous Driving}

\author[1]{Yuzhi Wu}
\ead{yuzhiwu1105@mail.ustc.edu.cn} 

\author[1,2]{Li Xiao}
\ead{xiaoli11@ustc.edu.cn}
\cormark[1] 

\author[3]{Jun Liu}
\ead{junliu@ustc.edu.cn}

\author[3]{Guangfeng Jiang}
\ead{jgf1998@mail.ustc.edu.cn} 

\author[4]{Xiang-Gen Xia}
\ead{xxia@ee.udel.edu}

\address[1]{MoE Key Laboratory of Brain-Inspired Intelligence Perception and Cognition, University of Science and Technology of China, Hefei 230052, China}
\address[2]{Institute of Artificial Intelligence, Hefei Comprehensive National Science Center, Hefei 230088, China}
\address[3]{Department of Electronic Engineering and Information Science, University of Science and Technology of China, Hefei 230027, China}
\address[4]{Department of Electrical and Computer Engineering, University of Delaware, Newark, DE 19716, USA}

\cortext[1]{Corresponding author}

\begin{abstract}
The emerging 4D millimeter-wave radar, measuring the range, azimuth, elevation, and Doppler velocity of objects, is recognized for its cost-effectiveness and robustness in autonomous driving. Nevertheless, its point clouds exhibit significant sparsity and noise, restricting its standalone application in 3D object detection. Recent 4D radar-camera fusion methods have provided effective perception. Most existing approaches, however, adopt explicit Bird's-Eye-View fusion paradigms originally designed for LiDAR-camera fusion, neglecting radar's inherent drawbacks. Specifically, they overlook the sparse and incomplete geometry of radar point clouds and restrict fusion to coarse scene-level integration. To address these problems, we propose MLF-4DRCNet, a novel two-stage framework for 3D object detection via multi-level fusion of 4D radar and camera images. Our model incorporates the point-, scene-, and proposal-level multi-modal information, enabling comprehensive feature representation. It comprises three crucial components: the Enhanced Radar Point Encoder (ERPE) module, the Hierarchical Scene Fusion Pooling (HSFP) module, and the Proposal-Level Fusion Enhancement (PLFE) module. Operating at the point-level, ERPE densities radar point clouds with 2D image instances and encodes them into voxels via the proposed Triple-Attention Voxel Feature Encoder. HSFP dynamically integrates multi-scale voxel features with 2D image features using deformable attention to capture scene context and adopts pooling to the fused features. PLFE refines region proposals by fusing image features, and further integrates with the pooled features from HSFP. Experimental results on the View-of-Delft (VoD) and TJ4DRadSet datasets demonstrate that MLF-4DRCNet achieves the state-of-the-art performance. Notably, it attains performance comparable to LiDAR-based models on the VoD dataset.
\end{abstract}


%\item Considering the characteristics of 4D radar point clouds, we propose a two-stage framework for 3D object detection via 4D radar-camera fusion, called MLF-4DRCNet. It fully exploits the complementary information from both radar and camera to enable hierarchical cross-modal interactions.
%\item MLF-4DRCNet comprises three key components, i.e., the Enhanced Radar Point Encoder module, the Hierarchical Scene Fusion Pooling module, and the Proposal-Level Fusion Enhancement module, which fuse 4D radar point clouds and camera images at the point, scene, and proposal levels, respectively.
%\item Experimental results on the VoD and TJ4DRadset datasets demonstrate the superiority of our proposed MLF-4DRCNet for 3D object detection. Remarkably, MLF-4DRCNet performs on par with some LiDAR-based models on the VoD dataset. 

% Keywords

\maketitle

% Main text
\section{Introduction}

Accurate 3D object detection constitutes the foundation for autonomous driving systems, which enables precise localization and classification of critical objects such as vehicles and pedestrians in 3D space \cite{fernandes2021point}. Existing high-precision 3D object detection algorithms typically remain dependent on LiDAR point clouds. Despite the fact that LiDAR sensors offer highly accurate geometric sensing, the high cost impedes their large-scale deployment and moreover the performance suffers from severe degradation under adverse weather conditions \cite{dreissig2023survey}. To address these constraints, millimeter-wave radar (hereinafter referred to simply as radar) has garnered considerable attention in the field of autonomous driving. The radar sensor allows superior distance measurement and velocity estimation while maintaining robust performance across diverse weather conditions \cite{zhou2022towards}. In addition, the emerging 4D radar technology, which measures range, azimuth, elevation, and Doppler velocity, effectively resolves the elevation limitation in traditional 3D radar, resulting in relatively dense 3D point clouds that spatially resemble LiDAR point clouds \cite{zhou2022towards}. This advancement mitigates radar point cloud sparsity while providing high-resolution sensing, positioning 4D radar as a viable option in autonomous driving perception systems.

\begin{figure}
    \centering
    \includegraphics[width=0.95\textwidth]{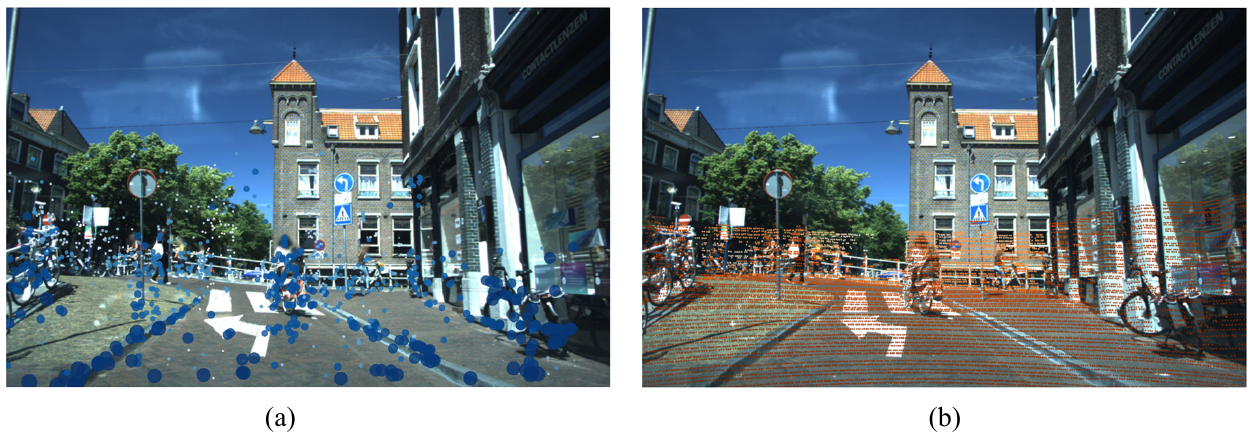}
    \caption{(a) 4D radar point clouds accumulated from adjacent consecutive five frames, (b) 64-line LiDAR point clouds in one frame.} \label{figure1}
\end{figure}

Several methods relying on 4D radar point clouds have been developed for 3D object detection \cite{palffy2022multi, xu2021rpfa, liu2023smurf}. Although technically feasible, they exhibit a significant performance gap compared to LiDAR-based approaches in detection accuracy. This performance gap mainly stems from two inherent limitations of radar point clouds, i.e., sparsity and noise. While 4D radar point clouds exhibit certain structural similarities to LiDAR point clouds, they remain comparatively sparse and noisy due to fundamental constraints including hardware-limited angular resolution and multipath effects, as illustrated by Fig. \ref{figure1} in a representative scenario from the View-of-Delft (VoD) dataset \cite{palffy2022multi}. From Fig. \ref{figure1}, accumulated radar point clouds over five frames yield merely 2,000 points, whereas a single frame from a 64-line LiDAR generates over 15,000 points, thus starkly demonstrating higher sparsity of 4D radar point clouds. To bridge the performance gap, a practical and promising strategy is to complement 4D radar with other sensors. Cameras serve as ideal fusion partners by delivering dense semantic information and rich textural details that compensate for radar sparsity. To this end, advancements in 4D radar-camera fusion have been made \cite{kim2023crn, zheng2023rcfusion,zhang2024tl,gu2025hgsfusion, bai2024sgdet3d, liu2025mssf, bi2025dual}. However, most existing 4D radar-camera fusion approaches are based on Bird's-Eye-View (BEV) fusion paradigms originally designed for LiDAR-camera fusion, where radar and image features are transformed into a unified BEV perspective through view transformation for cross-modal interaction. To enhance image features, the geometric cues from radar data are further leveraged to improve depth estimation \cite{zheng2023rcfusion, bai2024sgdet3d}. Critically, they overlook the fundamental characteristics of radar point clouds. Unlike LiDAR point clouds, radar point clouds are typically sparse and incomplete with inherent noise \cite{liu2025mssf}. Using geometric information from such radar data to guide monocular depth estimation frequently introduces errors, thus degrading detection performance. Furthermore, these methods often focus on the entire scene-level fusion while neglecting local feature integration, especially in the regions surrounding potential targets.

To address the aforementioned issues, we propose a Multi-Level Fusion with 4D Radar and Camera Network (MLF-4DRCNet) for 3D object detection. It is a two-stage framework consisting of three crucial components, i.e., the Enhanced Radar Point Encoder (ERPE) module, the Hierarchical Scene Fusion Pooling (HSFP) module, and the Proposal-Level Fusion Enhancement (PLFE) module. Specifically, the ERPE module densifies point clouds using 2D image instances, within which we introduce the Triple Attention Voxel Feature Encoder (TA-VFE) to encode the densified radar point clouds into voxels. A 3D backbone network subsequently abstracts these voxels into multi-scale 3D voxel feature volumes, while a 2D backbone and Region Proposal Network (RPN) generate high-quality region proposals. The HSFP module aims to capture scene-level features by leveraging deformable attention mechanisms \cite{zhu2020deformable} to dynamically integrate multi-scale 3D voxel features with 2D image features, avoiding explicit BEV transformation. This module also enables the generation of pooled proposal features essential for the following PLFE module. In PLFE, proposals from the RPN are first fused with image features via deformable attention (consistent with HSFP), then further integrated with the pooled proposal features derived from HSFP, yielding enhanced representations optimized for 3D object detection.

In summary, the proposed MLF-4DRCNet advances radar-camera 3D object detection through multi-level fusion: ERPE operates at the point level, HSFP at the scene level, and PLFE at the proposal level. This hierarchical structure for object detection ensures comprehensive feature representation, leading to improved detection results. Extensive experiments on the VoD and TJ4DRadset \cite{zheng2022tj4dradset} datasets demonstrate that our MLF-4DRCNet achieves the state-of-the-art (SOTA) performance. Notably, our model even achieves performance comparable to that of LiDAR-based models on the VoD dataset.

%The rest of this paper is structured as follows. Section II reviews single-modality and multi-modality fusion approaches for 3D object detection. Section III details the architecture of the proposed MLF-4DRCNet. Section IV presents experimental results evaluating MLF-4DRCNet on the VoD and TJ4DRadSet datasets, highlighting its effectiveness for 3D object detection. Finally, conclusion and future work are discussed in Section V.

\section{Related Work}
\subsection{LiDAR-based and Radar-based 3D Object Detection}
Based on the point cloud representation in LiDAR data processing, LiDAR-based 3D object detection methods are typically categorized as point-based, pillar-based, and voxel-based approaches. However, despite the effectiveness, LiDAR sensors suffer from two major drawbacks: high cost and degraded performance in adverse weather conditions. Conversely, radar, particularly the emerging 4D radar, offers superior robustness under such challenging conditions and presents a lower-cost viable option, establishing its value for autonomous perception systems.

Currently, radar-based 3D object detection approaches mostly follow the design of LiDAR-based ones. However, due to the inherent sparsity and noise of radar point clouds, these approaches typically convert radar points into voxels or pillars to construct regular structures. For example, RPFA-Net \cite{xu2021rpfa} employs a pillar-based design incorporating self-attention mechanism to capture global features.
RadarPillarNet \cite{zheng2023rcfusion} independently encodes spatial, Doppler velocity, and radar cross section (RCS) features during pillar extraction to exploit critical radar attributes. 
SMURF \cite{liu2023smurf} incorporates additional density features into pillar-based backbone through kernel density estimation, further enhancing detection performance.
MVFAN \cite{yan2023mvfan} reweights foreground and background radar points and incorporates Doppler velocity and reflectivity data into the backbone for feature extraction.
MUFASA \cite{peng2024mufasa} adopts both BEV and cylindrical view transformation to enhance radar point cloud feature extraction.
RadarPillars \cite{musiat2024radarpillars} utilizes pillar-based attention with radial velocity decomposition and sparsity-adaptive scaling to expand receptive fields. 

Overall, while radar offers better robustness in adverse weather conditions with lower cost, radar-based 3D detectors fail to achieve the performance of LiDAR-based 3D detectors due to inherent sparsity and noise of radar points. This performance gap highlights the need for radar-camera fusion in 3D object detection.

\subsection{LiDAR-Camera Fusion for 3D Object Detection}
LiDAR-camera fusion methodologies for 3D object detection are basically classified into three paradigms based on the stage of multi-modal integration: early fusion, middle fusion, and late fusion. Early fusion approaches, such as PointPainting \cite{vora2020pointpainting}, directly embed raw image features into point clouds before feeding data into the network. While straightforward to implement, these approaches often lack deep feature-level interactions and exhibit high sensitivity to sensor calibrations. Late fusion approaches, such as SparseFusion\cite{xie2023sparsefusion}, combine outputs from independent modality-specific detectors, but suffer from limited cross-modal feature interaction during proposal generation stages, frequently leading to suboptimal performance.

Among the three fusion paradigms, middle fusion approaches have attracted increased attention for facilitating the feature-level interaction between LiDAR and image modalities. This paradigm further branches into explicit and implicit approaches. Explicit approaches leverage geometric projection techniques such as Lift-Splat-Shoot (LSS) to lift image features into 3D space. For example, BEVFusion \cite{liang2022bevfusion} fuses LiDAR and image features on a unified BEV space. Deepinteraction \cite{yang2022deepinteraction} further proposes a cross-modal fusion strategy on both BEV and perspective-view spaces, conserving modality-specific information. However, these methods often face challenges in preserving semantic richness during geometric lifting. In contract, implicit approaches instead employ attention mechanisms \cite{vaswani2017attention} to adaptively align and aggregate contextual image semantics. Pioneering work like BEVFormer \cite{li2024bevformer} and LoGoNet \cite{li2023logonet} uses transformer queries to extract surround-view image features, while FUTR3D \cite{chen2023futr3d} directly fuses multi-sensor features via attention operations. 

In summary, LiDAR-camera fusion provides a foundational basis for multi-modal 3D object detection research, where fusion strategies of point cloud representations can be naturally extended to 4D radar-camera systems.

\begin{figure*}
\centering
\includegraphics[scale=0.64]{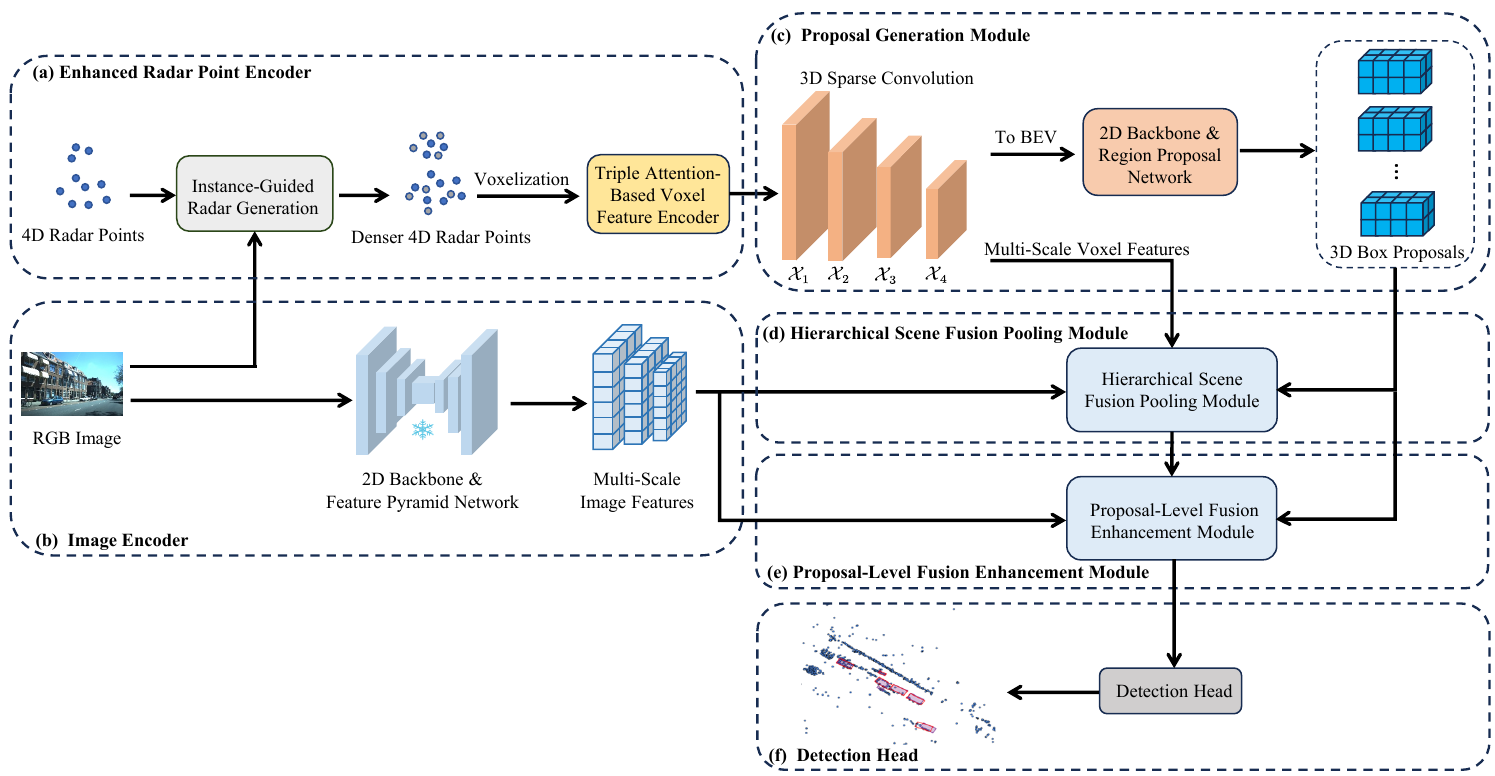}
\caption{The overall architecture of our MLF-4DRCNet. Raw radar points are first densified using 2D instance guidance from RGB images. The densified points are voxel-encoded via TA-VFE, and then processed by a 3D sparse backbone to generate multi-scale radar voxel features. After height compression, these features are sent to a 2D backbone and RPN to generate 3D proposals. Concurrently, 2D image features are extracted from RGB images. The HSFP module fuses multi-scale 3D radar voxel features with 2D image features to capture scene context and adopts pooling to the fused features. Finally, the PLFE module refines region proposals by fusing image features, and further integrates with the pooled features from the HSFP module. Finally, the aggregated proposal features are sent to the detection head to yield the detection results.}
\label{figure2}
\end{figure*}

\subsection{Radar-Camera Fusion for 3D Object Detection}
Following the LiDAR-camera fusion paradigm, several radar-camera fusion methods implement explicit fusion through cross-modal feature interaction in the BEV perspective. RCFusion \cite{zheng2023rcfusion} employs orthographic feature transformation for lifting image features, which are subsequently fused with radar BEV features through attention mechanisms. These approaches \cite{kim2023crn,lin2024rcbevdet} further utilize radar occupancy maps to guide the transformation of image features into the BEV perspective. SGDet3D \cite{bai2024sgdet3d} incorporates bidirectional feature interaction during feature extraction, leveraging radar geometry information to assist depth estimation while concurrently utilizing image semantic information to enhance radar features. However, unlike LiDAR point clouds, radar point clouds are typically sparse and incomplete with inherent noise. Using geometric information from such radar data to guide monocular depth estimation frequently introduces errors, resulting in suboptimal detection performance. To mitigate the sparsity, HGSFusion \cite{gu2025hgsfusion} first densifies radar point clouds using image information, while simultaneously converting both radar point clouds and images into the BEV perspective for fusion. Nevertheless, it fails to effectively utilize features from generated virtual points and distinguish their importance relative to raw radar points.

On the other hand, some radar-camera fusion approaches employ implicit fusion paradigms, leveraging cross-attention mechanisms to aggregate image information. For instance, CRAFT \cite{kim2023craft} refines image proposals using radar measurements through a cross-attention based spatio-contextual transformer. MSSF \cite{liu2025mssf} designs a fusion block leveraging deformable cross-attention mechanisms to dynamically interact point cloud features with image features via learnable offset matrices. However, these approaches primarily integrate radar point cloud and image features at a global scene level, leaving out the fusion of local information. In this study, our MLF-4DRCNet explores radar-camera fusion from both global scene-level and local proposal-level to thoroughly facilitate interaction between radar and image features.

\section{Methodology}
\subsection{Overall Architecture}
MLF-4DRCNet is a two-stage network for 3D object detection with 4D radar-camera fusion. As illustrated in Fig. \ref{figure2}, its overall architecture includes six components: the ERPE module, the image encoder, the proposal generation module, the HSFP module, the PLFE module, and a detection head. 

\begin{enumerate}
\item In the ERPE module, the raw radar point clouds are first densified using 2D instance guidance from RGB images. These denser radar points are then voxelized and encoded by the proposed TA-VFE.
\item The image encoder is used to extract 2D image features from the corresponding RGB images.
\item The proposal generation module processes the encoded features from TA-VFE using a 3D sparse convolutional backbone to yield multi-scale radar voxel features. After height compression, these features are passed through a 2D backbone and RPN to generate 3D region proposals.
\item At the scene-level, HSFP dynamically integrates multi-scale 3D voxel features with 2D image features using deformable attention to capture scene context, subsequently applying pooling to the fused features.
\item At the proposal-level, PLFE refines region proposals by fusing image features, and further integrates with the pooled features from HSFP. 
\item Finally, the detection head uses the refined proposal features from PLFE for bounding box refinement and confidence prediction.
\end{enumerate}

Details of each component of MLF-4DRCNet are presented in the following subsections.

\subsection{Enhanced Radar Point Encoder (ERPE)}
As a key component of MLF-4DRCNet, ERPE utilizes image instance segmentation results to perform point cloud densification, followed by voxel-based encoding of the augmented point clouds. Specifically, we first employ the point cloud densification method in \cite{gu2025hgsfusion, yin2021multimodal}, which operates on the Project-Sample-Reproject paradigm to generate virtual points using image instance segmentation results. To further enhance discrimination between raw and virtual points while fully leveraging virtual point information, we then propose TA-VFE to encode the hybrid point clouds.
\subsubsection{Instance Guided Radar Generation}
Suppose that we have raw radar points $
\mathcal{P}_{raw}=\{\mathbf{p}_i=\left( x_i,y_i,z_i,\mathbf{o}_i \right) \}_{i=1}^{N_{raw}}$, wherein $N_{raw}$ is the number of raw radar points, and $\left( x_i,y_i,z_i \right)$ and $\mathbf{o}_i$ represent the 3D location and other radar characteristics such as RCS and velocity of the $i$-th radar point, respectively. Given the 2D image corresponding to the current radar point cloud frame, a pretrained instance segmentation network \cite{cheng2022masked} is used to extract instances $\mathcal{M}=\left\{ \mathbf{m}_j,\ \mathbf{e}_j \right\} _{j=1}^{N_{mask}}$ from the image, where $
N_{mask}$ is the number of instances, $
\mathbf{m}_j
$ represents the $j$-th instance, and $\mathbf{e}_j$ is its associated binary labels. In this subsection, we use the raw radar points $
\mathcal{P}_{raw}
$ and 2D instance segmentation results $\mathcal{M}$ to generate virtual radar points.

Firstly, we project each raw radar point onto the corresponding 2D image based on the camera intrinsic matrix $\text{T}_{\text{intr}}\in \mathbb{R}^{3\times 4}$ and radar-to-camera coordinate transformation matrix $\text{T}_{\text{r2c}}\in \mathbb{R}^{4\times 4}$. Specifically, for the $i$-th radar point $\mathbf{p}_i=\left( x_i,y_i,z_i,\mathbf{o}_i \right)$, we leverage $\text{T}_{\text{intr}}$ and $\text{T}_{\text{r2c}}$ to transform its spatial coordinates into the image coordinate system, i.e.,
\begin{equation}\label{equation1}
\left[ \begin{array}{c}
	u_i\cdot d_i\\
	v_i\cdot d_i\\
	d_i\\
\end{array} \right] =\text{T}_{\text{intr}}\cdot \text{T}_{\text{r2c}}\cdot \left[ \begin{array}{c}
	x_i\\
	y_i\\
	z_i\\
	1\\
\end{array} \right] \ ,
\end{equation}
where $\left[ u_i,v_i \right]$ represents the pixel coordinates, and $d_i$ is the corresponding depth. Thus, we obtain $
\mathbf{p}_{img,i}=\left( u_i,v_i,d_i,\mathbf{o}_i \right) 
$ as the feature of $\mathbf{p}_i$ in the image coordinate system. Then we compare the locations of all projected points and all instances. For the $j$-th instance $\mathbf{m}_j$, we exclusively consider the projected points that fall within $\mathbf{m}_j$, i.e., foreground points, to construct the spatially filtered set 
\begin{equation}\label{equation0}
\mathcal{Q}_j=\left\{ \mathbf{p}_{fore,i}=\left( u_i,v_i,d_i,\mathbf{o}_i,\mathbf{e}_j \right) |\left( u_i,v_i \right) \in \mathbf{m}_j \right\} .
\end{equation}
These sets serve as the basis for generating virtual points.

Considering the characteristics of radar points, the probability of sampling virtual points should be higher in areas adjacent to foreground points than in other areas, and this probability increases as the distance to the foreground points decreases. Therefore, when sampling virtual points for each instance, we apply Gaussian sampling in regions near foreground points, whereas uniform sampling is utilized in other regions of the instance. Specifically, for a given instance $\mathbf{m}_j$, we denote its foreground point set as $\mathcal{Q}_j$. For any foreground point $\mathbf{p}_{fore,i}$ in $\mathcal{Q}_j$, its adjacent area within $\mathbf{m}_j$ is defined as
\begin{equation}\label{equation2}
\mathcal{R}_i\left( u,v \right) =\left\{ \left( u,v \right) \in \mathbf{m}_j|\left( u-u_i \right) ^2+\left( v-v_i \right) ^2<r^2 \right\},
\end{equation}
where $r$ represents the radius of the area. Within $
\mathcal{R}_i\left( u,v \right) 
$, the sampling points follow a Gaussian distribution with mean $
\left[ u_i,v_i \right] ^{\text{T}}
$ and covariance matrix $
\operatorname{diag}\left( \sigma _{1}^{2},\ \sigma _{2}^{2} \right)$. Assuming that there are $N_j$ foreground points on instance $\mathbf{m}_j$, we define $\mathcal{R}\left( u,v \right) =\underset{i=1}{\overset{N_j}{\cup}}\mathcal{R}_i\left( u,v \right)$ as the union of the neighborhoods associated with all foreground points, and sample a fixed number $
\tau 
$ of points in $\mathcal{R}\left( u,v \right)$. Additionally, in other regions of instance $\mathbf{m}_j$, we uniformly sample $\tau$ points owing to the lack of prior information. Consequently, all sampled points on instance $\mathbf{m}_j$ can be represented as
\begin{equation}\label{equation4}
\mathcal{S}_j=\{\{\mathbf{s}_{gauss,i}=\left( u_i,v_i \right) \}_{i=1}^{\tau},\ \left\{ \mathbf{s}_{uni,i}=\left( u_i,v_i \right) \right\} _{i=1}^{\tau}\},
\end{equation}
where $\mathbf{s}_{gauss,i}$ and $\mathbf{s}_{uni,i}$ represent the points sampled through Gaussian sampling and uniform sampling, respectively. Next, we perform depth estimation on the sampled points to reproject them back to 3D space.

For each Gaussian sampled point $\mathbf{s}_{gauss,i}$ on instance $\mathbf{m}_j$, we calculate the spatial distances from it to each foreground point in $\mathcal{Q}_j$, and retrieve the depth estimate and other feature estimates from its nearest neighbor in the set. This progress can be expressed as
\begin{equation}\label{equation5}
\left( d_i,\mathbf{o}_i,\mathbf{e}_j \right) =\argmin_{\left( d_k,\mathbf{o}_k,\mathbf{e}_j \right)}\lVert \mathbf{p}_{fore,k}^{'}-\mathbf{s}_{gauss,i} \rVert ,
\end{equation}
where $\mathbf{p}_{fore,k}^{'}=\left( u_k,v_k \right)$ is the spatial coordinate of the foreground point $\mathbf{p}_{fore,k}$. The estimated value $\left( d_i,\mathbf{o}_i,\mathbf{e}_j \right)$ is assigned to $\mathbf{s}_{gauss,i}$, resulting in $
\mathbf{s}_{gauss,i}^{'}=\left( u_i,v_i,d_i,\mathbf{o}_i,\mathbf{e}_j \right) 
$. Then we leverage the inverse of (\ref{equation1}), reproject the generated point back to the radar coordinate, and obtain the corresponding virtual 3D point.

Uniformly sampled points are processed similarly to Gaussian sampled points, except that each uniformly sampled point is assigned the corresponding estimates of its four closest foreground points following \cite{gu2025hgsfusion}. This procedure yields four 3D virtual points per uniformly sampled point. This is because that Gaussian sampled points concentrate in high-confidence regions, where depth consistency is strong. Single-neighbor matching preserves precision while minimizing computation. Uniformly sampled points cover noisy peripheries where depth varies. Multi-neighbor matching enhances surface coverage through spatial averaging, mitigating edge uncertainty \cite{li2026dynamic}.

In summary, the generated 3D virtual points can be presented as $
\mathcal{P}_{virtual}=\{\mathbf{p}_{vir,i}=\left( x_i,y_i,z_i,\mathbf{o}_i,\mathbf{e}_i \right) \}_{i=1}^{N_{virtual}}
$, where $N_{virtual}$ is the number of virtual points. 
To align the feature dimensions of the raw radar points with those of the generated virtual points, we apply unit padding to the features of the raw points. This process yields the augmented set
$
\hat{\mathcal{P}}_{raw}=\{\mathbf{p}_i=\left( x_i,y_i,z_i,\mathbf{o}_i,1\mathbf{s} \right) \}_{i=1}^{N_{raw}}
$, where $1\mathbf{s}$ represents the unit-padded feature.
Furthermore, to explicitly distinguish between point types, we augment the feature vectors of both sets using one-hot encoding vectors $
\{\mathbf{r}_i\}$ to represent their types, resulting in the final two point sets:
\begin{equation}\label{equation6}
\tilde{\mathcal{P}}_{raw}=\{\mathbf{p}_i=\left( x_i,y_i,z_i,\mathbf{o}_i,1\mathbf{s,\ r}_i \right) \}_{i=1}^{N_{raw}}\  \text{and}
\end{equation}
\begin{equation}\label{equation7}
\tilde{\mathcal{P}}_{virtual}=\{\mathbf{p}_{vir,i}=\left( x_i,y_i,z_i,\mathbf{o}_i,\mathbf{e}_i,\mathbf{r}_i \right) \}_{i=1}^{N_{virtual}}.
\end{equation}

\begin{figure}[pos=b]
\centering
\includegraphics[scale=0.78]{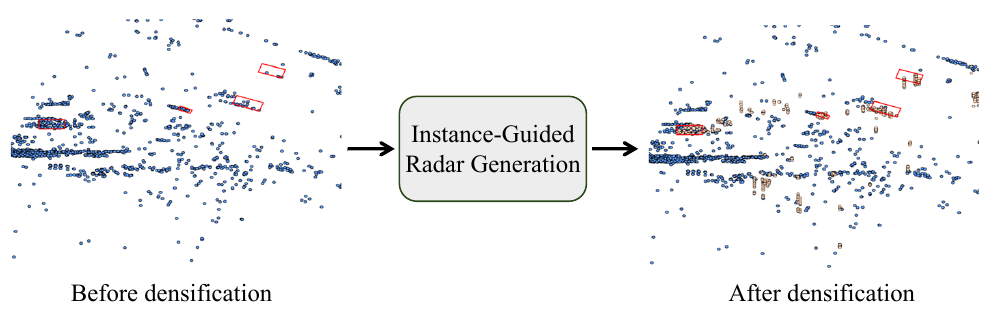}
\caption{Comparative BEV visualization of radar point cloud before and after densification. Red boxes denote the ground truth regions of objects to be detected (cars, pedestrians, and cyclists). Raw radar points are shown in blue and generated virtual points are shown in yellow.}
\label{figure3}
\end{figure}

As shown in Fig. \ref{figure3}, we increase the density of the radar point clouds in object areas to facilitate detection. The combined set of hybrid points, $\mathcal{P}=\left\{ \tilde{\mathcal{P}}_{raw},\tilde{\mathcal{P}}_{virtual} \right\} 
$, is then fed into TA-VFE for feature encoding.

\subsubsection{Triple Attention Voxel Feature Encoder (TA-VFE)}
Although the generated virtual points alleviate the sparsity of raw radar points, their inherent geometric discrepancies with raw points introduce distribution shifts. Besides, virtual points carry additional semantic features which also induce biases in feature channels. To address these challenges, we propose TA-VFE, which integrates the Triple Attention mechanism \cite{liu2020tanet} into voxel feature encoding, processing hybrid points $\mathcal{P}$ through point-wise, channel-wise, and voxel-wise attention.

\begin{figure}
\centering
\includegraphics[scale=0.6]{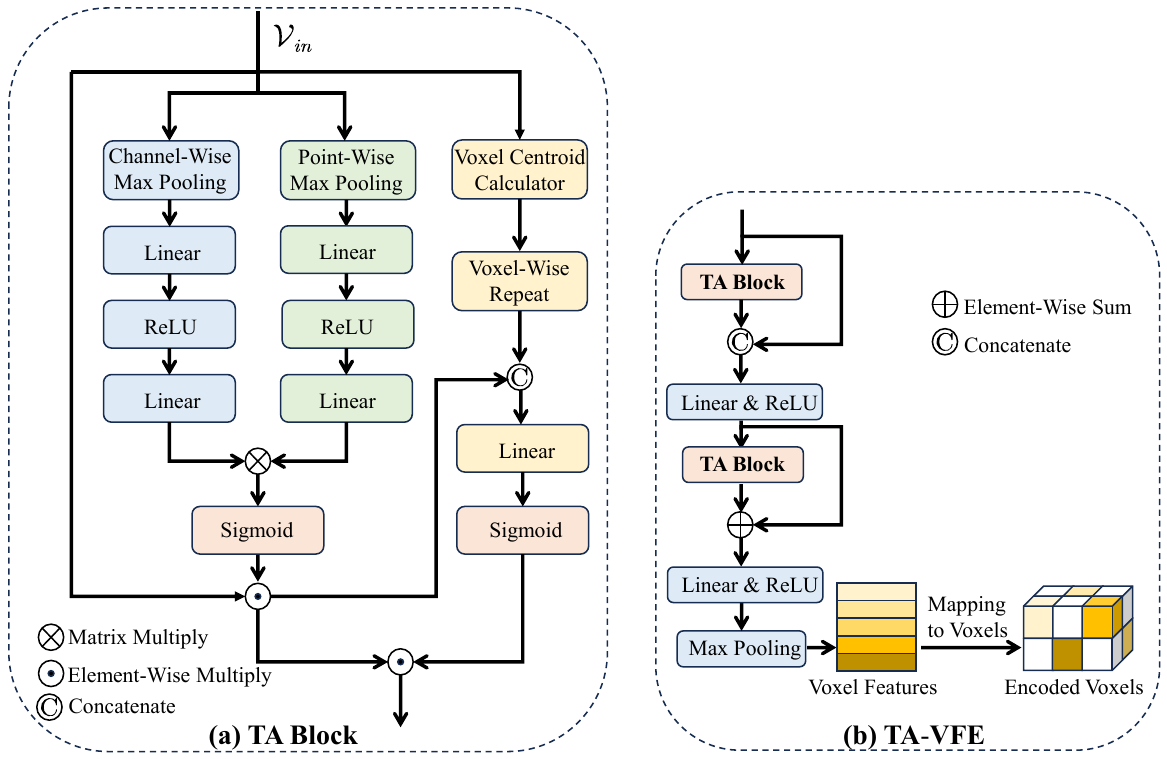}
\caption{(a) Illustration of the TA block. (b) Overall architecture of TA-VFE, which stacks two TA blocks.}
\label{figure4}
\end{figure}

The overall architecture of TA-VFE is illustrated in Fig. \ref{figure4}, primarily comprising two Triple Attention (TA) blocks. The detailed structure of the TA block is depicted in Fig. \ref{figure4}(a). Specifically, the hybrid points are first divided into small voxels with spatial resolution of $
L\times H\times W$, and encoded as a voxel set $
\mathcal{V}_{in}=\{\mathbf{P}_{in},\mathbf{C}_V\}
$. Here, $
\mathbf{P}_{in}\in \mathbb{R}^{N_V\times N_P\times C_{in}}$ stores the information of points within each non-empty voxel, where $N_V$ is the number of non-empty voxels, $N_P$ denotes the predefined maximum number of points per voxel, and $C_{in}$ represents the feature dimension per point. Voxels containing fewer than $N_P$ points are padded with zeros. The voxel indices are stored in $
\mathbf{C}_V\in \mathbb{R}^{N_V\times 3}$.

Let $\mathcal{V}_{in}^{k}$ denote the $k$-th non-empty voxel with feature $
\mathbf{P}_{in}^{k}\in \mathbb{R}^{N_p\times C_{in}}$.  The intra-voxel centroid coordinate $
\mathbf{c}^k\in \mathbb{R}^3$ is calculated by
\begin{equation}\label{equation8}
\mathbf{c}^k=\frac{1}{\left| \mathcal{V}_{in}^{k} \right|}\sum_{j=1}^{\left| \mathcal{V}_{in}^{k} \right|}{\mathbf{p}_{j}^{k}},
\end{equation}
where $\left| \mathcal{V}_{in}^{k} \right|$ is the number of actual points in $\mathcal{V}_{in}^{k}$, and $
\mathbf{p}_{j}^{k}\in \mathbb{R}^3$ denotes the spatial coordinate of the $j$-th actual point in $\mathcal{V}_{in}^{k}$, $
j=1,\cdots,\left| \mathcal{V}_{in}^{k} \right|$. We leverage the centroid to enable spatial context augmentation. According to \cite{lang2019pointpillars}, each point in $\mathcal{V}_{in}^{k}$ has its feature enhanced by appending its offsets to both the voxel geometric center and intra-voxel centroid. This yields an augmented feature $
\mathbf{\hat{P}}_{in}^{k}\in \mathbb{R}^{N_p\times C_{in}^{'}}$, where $C_{in}^{'}=C_{in}+6$. These incorporated geometric attributes provide foundational positional cues for subsequent feature extraction.

Then max-pooling operations are applied to the augmented feature $\mathbf{\hat{P}}_{in}^{k}$ along the point dimension and channel dimension, respectively, yielding point-wise pooled feature $
\mathbf{\hat{P}}_{pw}^{k}\in \mathbb{R}^{N_p\times 1}$ and channel-wise pooled feature $
\mathbf{\hat{P}}_{cw}^{k}\in \mathbb{R}^{1\times C_{in}^{'}}$. The attention weights are computed as
\begin{equation}\label{equation9}
\mathbf{W}_{pw}^{k}=\text{LN}\left( \text{$\delta$}\left( \text{LN}\left( \mathbf{\hat{P}}_{pw}^{k} \right) \right) \right)\  \text{and}
\end{equation}
\begin{equation}\label{equation10}
\mathbf{W}_{cw}^{k}=\text{LN}\left( \text{$\delta$}\left( \text{LN}\left( \mathbf{\hat{P}}_{cw}^{k} \right) \right) \right) ,
\end{equation}
where $\text{LN}$ denotes a linear layer, and $\delta$ represents the ReLU activation function. $\mathbf{W}_{pw}^{k}\in \mathbb{R}^{N_p\times 1}$ is the the point-wise attention weight, quantifying spatial relationships of points within voxel $
\mathcal{V}_{in}^{k}$. $\mathbf{W}_{cw}^{k}\in \mathbb{R}^{1\times C_{in}^{'}}$ is the channel-wise attention weight, weighting feature channel contributions for $
\mathcal{V}_{in}^{k}$. The combined attention weight matrix $
\mathbf{M}_{pcw}^{k}\in \mathbb{R}^{N_p\times C_{in}^{'}}$ is obtained through
\begin{equation}\label{equation11}
\mathbf{M}_{pcw}^{k}=\sigma \left( \mathbf{W}_{pw}^{k}\times \mathbf{W}_{cw}^{k} \right) ,
\end{equation}
where $\sigma$ denotes the sigmoid function. The weighted voxel feature $
\mathbf{P}_{pcw}^{k}\in \mathbb{R}^{N_p\times C_{in}^{'}}$ is then computed via element-wise multiplication, i.e.,
\begin{equation}\label{equation12}
\mathbf{P}_{pcw}^{k}=\mathbf{M}_{pcw}^{k}\odot \mathbf{\hat{P}}_{in}^{k}.
\end{equation}

Building upon the aforementioned point-wise and voxel-wise attention mechanisms, we further implement voxel-wise attention to evaluate the importance of individual voxels within the global grid. For the $k$-th non-empty voxel with its updated feature $
\mathbf{P}_{pcw}^{k}$ and centroid coordinate $\mathbf{c}^k$, we first construct an enriched representation by concatenating the broadcasted centroid coordinate to each point's feature, i.e.,
\begin{equation}\label{equation13}
\mathbf{\hat{P}}_{pcw}^{k}=\left[ \mathbf{P}_{pcw}^{k},\ \mathbf{c}^k \right] \in \mathbb{R}^{N_p\times \left( C_{in}^{'}+3 \right)}.
\end{equation}

This representation then undergoes sequential dimensionality reduction through linear layers, first compressing the point-wise dimension and subsequently reducing the channel dimension to produce a scalar voxel-wise attention weight $
q^k\in \mathbb{R}$. The original voxel feature $\mathbf{P}_{pcw}^{k}$ is weighted by $
q^k$ to produce more robust feature, which can be formulated as
\begin{equation}\label{equation14}
\mathbf{P}_{out}^{k}=q^k\cdot \mathbf{P}_{pcw}^{k},
\end{equation}
where $\mathbf{P}_{out}^{k}\in \mathbb{R}^{N_p\times C_{in}^{'}}$ represents the attention-weighted feature of the $k$-th non-empty voxel through a TA block.

As illustrated in Fig. \ref{figure4}(b), TA-VFE stacks two TA blocks to progressively extract voxel features, followed by linear layers. Channel-wise max pooling is subsequently applied to aggregating point features within each non-empty voxel, producing the final voxel feature $\mathbf{P}_{final}\in \mathbb{R}^{N_V\times C_{out}}$, where $C_{out}$ denotes the output channel dimension for voxel feature extraction. In Fig. \ref{figure4}(b), different voxel features are visualized by different shades of yellow and spatially remapped to their original grid positions, forming the final output $
\mathcal{V}_{out}=\{\mathbf{P}_{final},\mathbf{C}_V\}$. Through these operations, we enhance key features of the hybrid points while suppressing irrelevant features introduced by virtual points, thus obtaining more robust voxel features.

\subsection{Image Encoder}
The image encoder extracts hierarchical features with rich semantic information from input images. Concretely, the RGB image $
\mathbf{I}$ is first encoded by a pretrained ResNet-101 \cite{he2016deep}. Features extracted from different receptive fields are then fused by a feature pyramid network \cite{lin2017feature}, yielding multi-scale outputs $
\mathbf{F}_{I,j}\in \mathbb{R}^{_{W_j\times H_j\times C_j}}
$, $j=1,\cdots ,n_I$, where $n_I$ is the number of scales, and $W_j$, $H_j$, $C_j$ denote the width, height and channel dimensions at the $j$-th scale, respectively.

\subsection{Proposal Generation Module}
The proposal generation module converts hybrid point cloud voxel features into high-quality 3D region proposals. Specifically, given the voxel-encoded inputs $\mathcal{V}_{out}=\{\mathbf{P}_{final},\mathbf{C}_V\}$ from TA-VFE, we use 3D sparse convolution networks \cite{yan2018second} to gradually transform $\mathcal{V}_{out}$ into feature volumes with 1×, 2×, 4×, and 8× downsampling rates, respectively. The output sparse features are denoted as $
\mathcal{X}_i=\{\mathbf{F}_{V,i},\ \mathbf{C}_{V,i}\},\ i=1,\cdots,4$, where $\mathbf{F}_{V,i}\in \mathbb{R}^{N_i\times C_i}$ represents the features of $N_i$ non-empty voxels generated by the $i$-th sparse convolutional layer, and $\mathbf{C}_{V,i}\in \mathbb{R}^{N_i\times 3}$ represents the coordinates of these voxels.

Following the anchor-based methods \cite{shi2020pv,deng2021voxel}, the 8× downsampled 3D voxel feature $\mathcal{X}_4$ is compressed along the height to produce BEV feature maps, followed by a 2D backbone network. This 2D backbone comprises a top-down feature extraction network containing two blocks of 3×3 convolutional layers, coupled with a multi-scale feature fusion network responsible for upsampling and concatenating the hierarchical features. Finally, the outputs of the 2D backbone are fed into parallel 1×1 convolutions, and then sent to RPN to generate 3D proposals $
\mathcal{B}=\{\mathbf{B}_1,\mathbf{B}_2,\cdots, \mathbf{B}_{N_{proposal}}\}$, where $N_{proposal}$ is the number of generated proposals.

\subsection{Hierarchical Scene Fusion Pooling Module (HSFP)}
\begin{figure}
\centering
\includegraphics[scale=0.4]{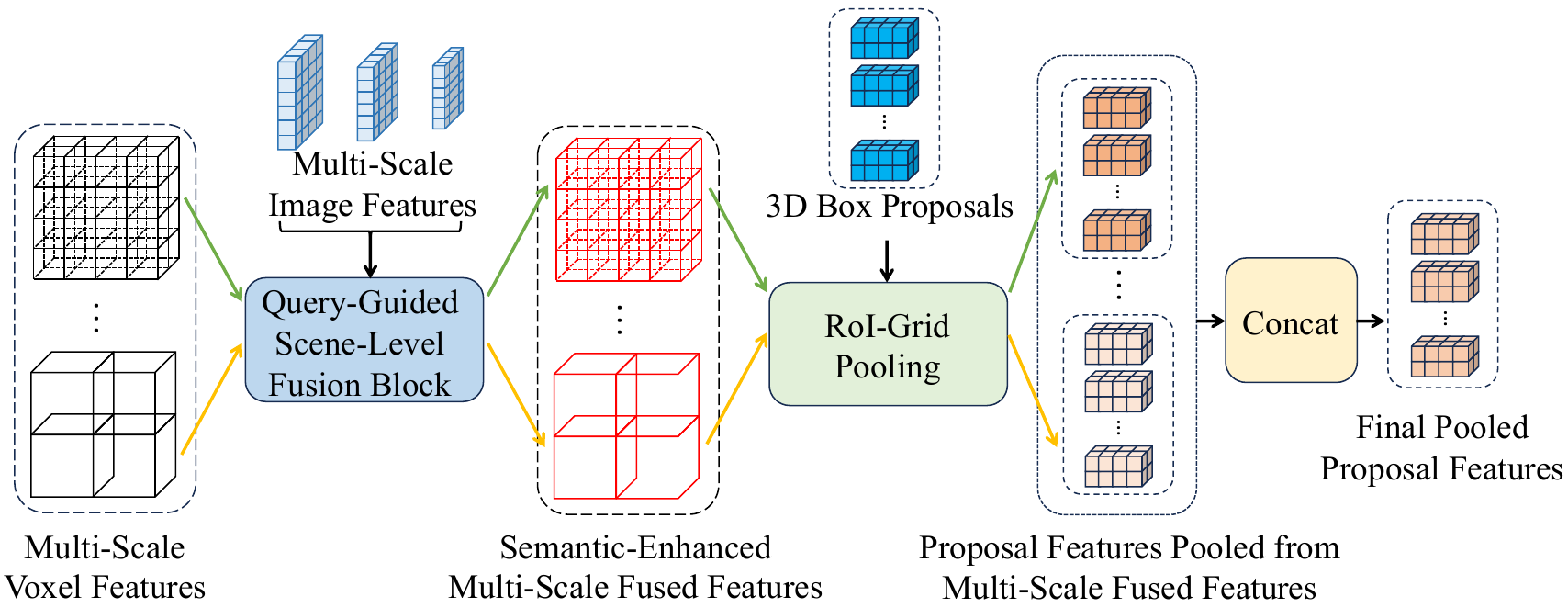}
\caption{Overview of the HSFP module. It individually fuses multi-scale 3D voxel features with image features, performs pooling on these fused features, and concatenates the pooled features to form the final representation.}
\label{figure5}
\end{figure}

As shown in Fig. \ref{figure5}, HSFP individually fuses multi-scale voxel features with 2D image features, performs pooling on these fused features, and concatenates the pooled features to form the final representation. The core unit is the Query-Guided Scene-Level Fusion (QGSLF) block, which leverages deformable attention mechanism \cite{zhu2020deformable} to enable dynamic cross-modal feature integration while avoiding explicit BEV transformation. Multi-scale voxel representations $\{\mathcal{X}_i\}_{i=1}^{4}$ are first extracted from the sparse convolutional layers in the proposal generation module, where each $
\mathcal{X}_i=\{\mathbf{F}_{3D,i},\ \mathbf{C}_{3D,i}\}
$ contains features $
\mathbf{F}_{3D,i}\in \mathbb{R}^{N_i\times C_i}$ and spatial coordinates $
\mathbf{C}_{3D,i}\in \mathbb{R}^{N_i\times 3}$ for $N_i$ non-empty voxels.

\begin{figure}
\centering
\includegraphics[scale=0.5]{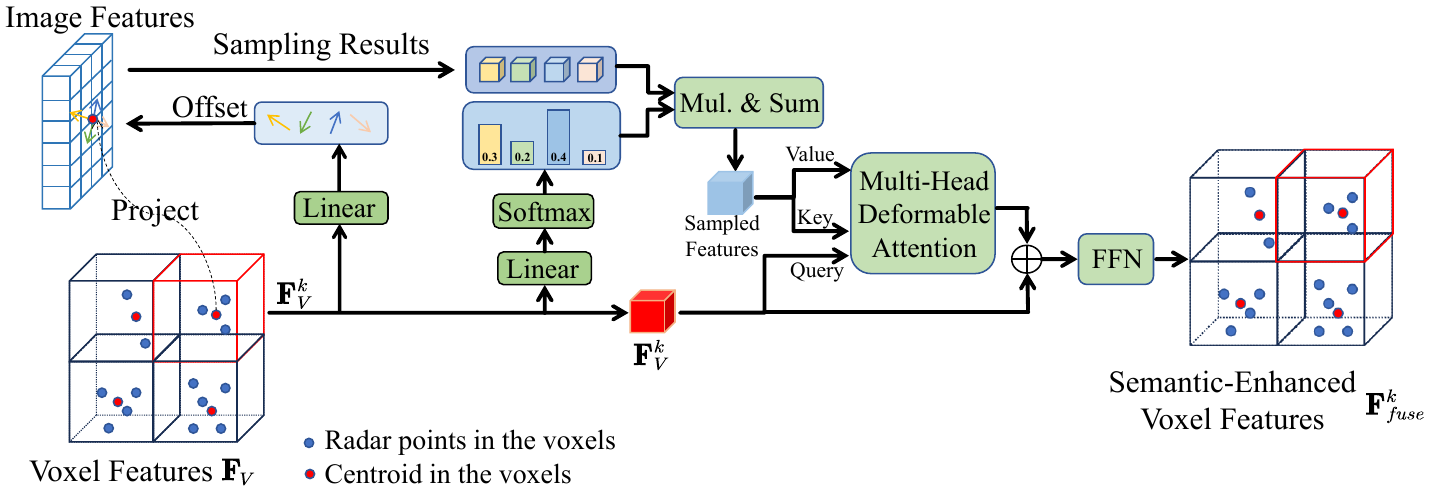}
\caption{Overview of the QGSLF block. For easy visualization, image features are shown in a single scale.}
\label{figure6}
\end{figure}

As illustrated in Fig. \ref{figure6}, each of these voxel features is individually fused with multi-scale image features via the QGSLF block. We detail the QGSLF block using a generic single-scale voxel representation $\mathcal{X}_i$ fused with multi-scale image features $
\left\{ \mathbf{F}_{I,j} \right\} _{j=1}^{n_I}$. Specifically, for the $k$-th non-empty voxel $\mathcal{X}_{i}^{k}$ with feature $\mathbf{F}_{V}^{k}\in \mathbb{R}^{C_i}$, we first calculate its centroid $
\mathbf{c}_{V}^{k}\in \mathbb{R}^3$ by averaging the spatial positions of all intra-voxel points, as in (\ref{equation8}). This centroid is then projected to 2D image coordinates via perspective transformation, i.e.,
\begin{equation}\label{equation15}
\left[ \begin{array}{c}
	u_k\cdot d_k\\
	v_k\cdot d_k\\
	d_k\\
\end{array} \right] =\text{T}_{\text{intr}}\cdot \text{T}_{\text{r2c}}\cdot \left[ \begin{array}{c}
	\mathbf{c}_{V}^{k}\\
	1\\
\end{array} \right] \ ,
\end{equation}
where $\left[ u_k,v_k \right]$ and $d_k$ denote the pixel coordinates and corresponding depth, respectively. Here, $\mathbf{c}_{img}^{k}=\left[ u_k,v_k \right] ^{\text{T}}$ denotes the projected centroid position of the $k$-th non-empty voxel. Via centroid projection, which provides geometric priors, we establish geometric correspondence that enhances cross-modal alignment \cite{liu2025mssf, li2023logonet}, effectively bridging the gap between voxel and image representations.


Referring to the multi-scale deformable attention mechanism \cite{zhu2020deformable}, we regard the voxel feature $\mathbf{F}_{V}^{k}$ as the query, with the projected centroid position $\mathbf{c}_{img}^{k}$ serving as the corresponding reference point. For multi-scale image features $
\left\{ \mathbf{F}_{I,j} \right\} _{j=1}^{n_I}$, we sample $n_s$ features from each feature map for one query through $M$ attention heads. Two parallel linear projection layers are employed to transform $\mathbf{F}_{V}^{k}$ to predict sampling parameters for each attention head $m\in \left\{ 1,\cdots,M \right\}$, each scale $j\in \left\{ 1,\cdots,n_I \right\}$, and each sampling point $
l\in \{1,\cdots,n_s\}$, i.e.,
\begin{equation}\label{equation16}
\bigtriangleup \mathbf{p}_{mjl}^{k}=\mathbf{W}_{offset}\mathbf{F}_{V}^{k},
\end{equation}
\begin{equation}\label{equation17}
a_{mjl}^{k}=\mathbf{W}_{attn}\mathbf{F}_{V}^{k},
\end{equation}
where $\mathbf{W}_{offset}\in \mathbb{R}^{2\times C_i}$ and $
\mathbf{W}_{attn}\in \mathbb{R}^{1\times C_i}$ are learnable weights, $
\bigtriangleup \mathbf{p}_{mjl}^{k}\in \mathbb{R}^2$ is the offset of the $l$-th sampling point relative to $\mathbf{c}_{img}^{k}$ in the $m$-th attention head at scale $j$, and $a_{mjl}^{k}\in \mathbb{R}$ denotes the corresponding attention weight. The scalar attention weight is then normalized over all scales and sampling points via the softmax operation, i.e.,
\begin{equation}\label{equation18}
\tilde{a}_{mjl}^{k}=\text{softmax} \left( a_{mjl}^{k} \right) =\frac{\exp \left( a_{mjl}^{k} \right)}{\sum_{j^{'}=1}^{n_I}{\sum_{l^{'}=1}^{n_s}{\exp \left( a_{mj^{'}l^{'}}^{k} \right)}}}\ .
\end{equation}
The sampling coordinates at each scale are then computed by applying offsets to the reference point, and image features are sampled via bilinear interpolation:
\begin{equation}\label{equation19}
\mathbf{p}_{mjl}^{k}=\mathbf{c}_{img}^{k}+\bigtriangleup \mathbf{p}_{mjl}^{k},
\end{equation}
\begin{equation}\label{equation20}
\mathbf{f}_{mjl}^{k}=\text{Sample}\left( \mathbf{F}_{I,j},\mathbf{p}_{mjl}^{k} \right) ,
\end{equation}
where the Sample operation is to extract $\mathbf{f}_{mjl}^{k}$ from the image feature $\mathbf{F}_{I,j}$ at coordinates $\mathbf{p}_{mjl}^{k}$ via bilinear interpolation. Subsequently, value projection and aggregation are performed across all scales, sampling points, and attention heads:
\begin{equation}\label{equation21}
\mathbf{\hat{F}}_{V}^{k}=\sum_{m=1}^M{\mathbf{W}_m}\sum_{j=1}^{n_I}{\sum_{l=1}^{n_s}{\tilde{a}_{mjl}^{k}\cdot \left( \mathbf{W}_{m}^{'}\mathbf{f}_{mjl}^{k} \right)}} ,
\end{equation}
where $\mathbf{W}_m$ and $\mathbf{W}_{m}^{'}$ are learnable weights, and $
\mathbf{\hat{F}}_{V}^{k}$ denotes the image-enhanced feature of the $k$-th non-empty voxel. The final fused voxel feature combines original and image-enhanced features as
\begin{equation}\label{equation22}
\mathbf{F}_{fuse}^{k}=\text{FFN}\left( \text{Concat}\left( \mathbf{F}_{V}^{k},\mathbf{\hat{F}}_{V}^{k} \right) \right) ,
\end{equation}
where FFN stands for a two-layer feed-forward network with LayerNorm and ReLU activation. This geometrically adaptive fusion mechanism enables the voxel feature to identify and integrate semantically relevant information from corresponding regions across all image scales. The process executes for all non-empty voxels in $\mathcal{X}_i$, generating semantically enhanced features $
\mathbf{F}_{scene}^{i}$, from which RoI grid pooling \cite{shi2020pv} extracts corresponding proposal features $
\mathbf{F}_{SLP}^{i}=\left[ \mathbf{F}_{SLP}^{i,1},\mathbf{F}_{SLP}^{i,2},\cdots,\mathbf{F}_{SLP}^{i,N_{propsoal}} \right]$ with 3D proposals $\mathcal{B}$. 

Notably, as shown in Fig. \ref{figure5}, the aforementioned pipeline individually performs scene-level fusion and pooling operations of multi-scale voxels and multi-scale image features. For the pooled features extracted at different scales, we concatenate them to form the final pooled features $\mathbf{F}_{SLP}$.

\subsection{Proposal-Level Fusion Enhancement Module (PLFE)}
\begin{figure}
\centering
\includegraphics[scale=0.52]{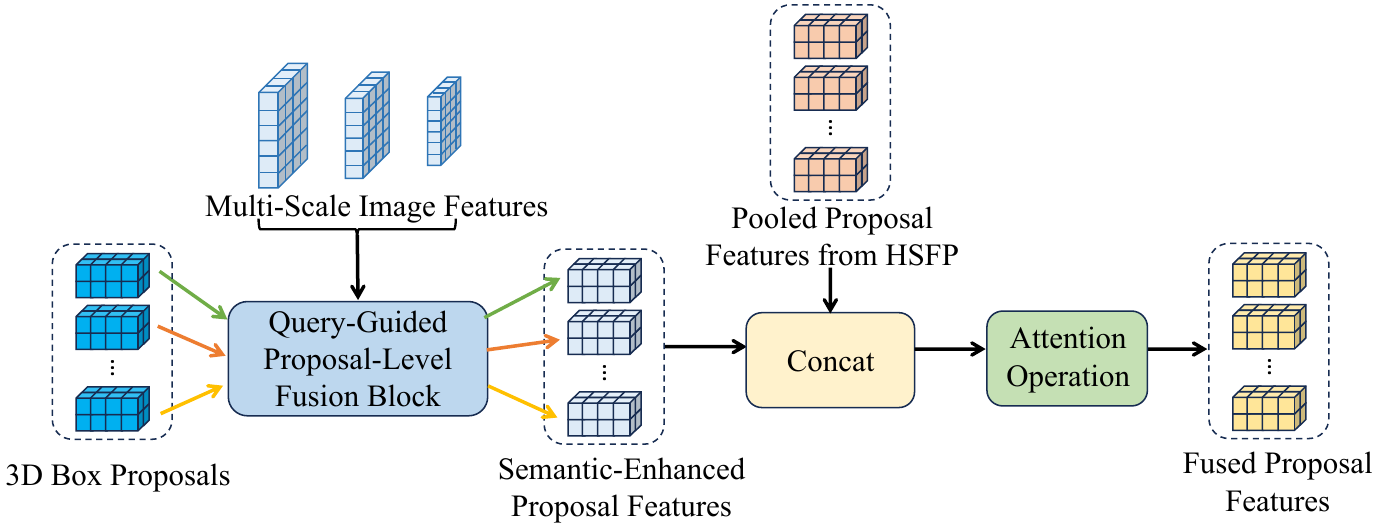}
\caption{Overview of the PLFE module. It performs proposal-level feature fusion with image features, followed by attention-based aggregation with semantically enhanced pooled features from the HSFP module to achieve comprehensive proposal refinement}
\label{figure7}
\end{figure}

The PLFE module is designed to mine multi-modal contextual information surrounding each proposal. As illustrated in Fig. \ref{figure7}, it performs proposal-level feature fusion by integrating intra-proposal geometric information with image features, followed by attention-based aggregation with semantically enhanced pooled features from HSFP to achieve comprehensive proposal refinement. 

\begin{figure}
\centering
\includegraphics[scale=0.48]{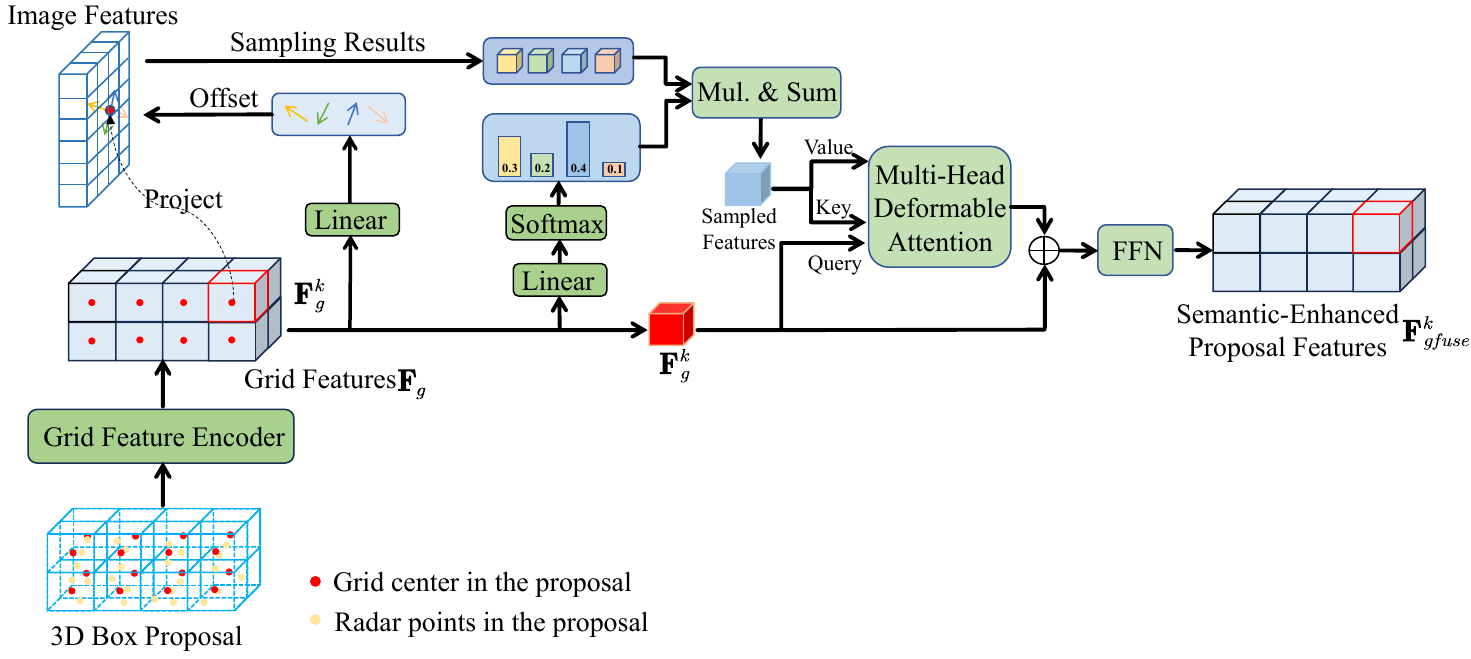}
\caption{Overview of the QGPLF block. For easy visualization, image features are shown in single scale.}
\label{figure8}
\end{figure}

Specifically, 3D proposals $\mathcal{B}$ and multi-scale image features $\left\{ \mathbf{F}_{I,j} \right\} _{j=1}^{n_I}$ are first fed into the Query-Guided Proposal-Level Fusion (QGPLF) block. Within QGPLF, we adopt the same design principle as QGSLF, utilizing deformable attention mechanisms to dynamically aggregate intra-proposal geometric information with corresponding image features. As shown in Fig. \ref{figure8}, mirroring the QGSLF's voxel-based methodology, we discretize each proposal into an $N_g=d_1\times d_2\times d_3$ grid and then perform feature encoding on these grid cells. Taking the fusion of an arbitrary proposal $
\mathbf{B}_i\in \mathcal{B}$ with multi-scale image features $
\left\{ \mathbf{F}_{I,j} \right\} _{j=1}^{n_I}$ as an example, for the $k$-th grid cell $G^k$ within $\mathbf{B}_i$, we calculate its geometric center $
\mathbf{t}_{grid}^{k}\in \mathbb{R}^3$ and centroid $
\mathbf{c}_{grid}^{k}\in \mathbb{R}^3$, then construct and encode the spatial representation through FFN, i.e.,
\begin{equation}\label{equation23}
\mathbf{f}_{grid}^{k}=\text{Concat}\left( \mathbf{t}_{grid}^{k},\ \mathbf{c}_{grid}^{k} \right) ,
\end{equation}
\begin{equation}\label{equation24}
\mathbf{F}_{g}^{k}=\text{FFN}\left( \mathbf{f}_{grid}^{k} \right) ,
\end{equation}
where $\mathbf{F}_{g}^{k}$ denotes the feature embedding of the grid cell $G^k$, serving as the query for deformable attention operation. This encoding preserves spatial relationships while compressing information for attention mechanisms.


Afterwards, the geometric center $\mathbf{t}_{grid}^{k}$ is projected onto the image using (\ref{equation15}), yielding the reference point $\mathbf{t}_{img}^{k}$ for sampling image features. For each query, $n_s$ features are sampled from each feature map of $
\left\{ \mathbf{F}_{I,j} \right\} _{j=1}^{n_I}$ through $M$  attention heads. The query $\mathbf{F}_{g}^{k}$ is processed through two linear layers to generate offsets $
\{\bigtriangleup \mathbf{\hat{p}}_{mjl}^{k}\in \mathbb{R}^2\}$ and weights $\{b_{mjl}^{k}\in \mathbb{R}\}$ for each head $
m\in \{1,\cdots,M\}$, each scale $j\in \{1,\cdots,n_I\}$, and each sampling point $l\in \{1,\cdots,n_s\}$. The attention weights are normalized across all scales and sampling points via the softmax operation, i.e.,
\begin{equation}\label{equation25}
\tilde{b}_{mjl}^{k}=\text{softmax} \left( b_{mjl}^{k} \right) =\frac{\exp \left( b_{mjl}^{k} \right)}{\sum_{j^{'}=1}^{n_I}{\sum_{l^{'}=1}^{n_s}{\exp \left( b_{mj^{'}l^{'}}^{k} \right)}}}\ .
\end{equation}

The corresponding sampled features $\mathbf{f}_{mjl}^{k}$ are obtained from each scale's feature map $\mathbf{F}_{I,j}$  using the  reference point $\mathbf{t}_{img}^{k}$ and its dynamic offsets $\bigtriangleup \mathbf{\hat{p}}_{mjl}^{k}$ with the same operations as those in (\ref{equation19}) and (\ref{equation20}). Then we obtain the image-enhanced grid feature $\mathbf{\hat{F}}_{g}^{k}$ by aggregating the sampled features using the normalized weights:
\begin{equation}\label{equation26}
\mathbf{\hat{F}}_{g}^{k}=\sum_{m=1}^M{\mathbf{\tilde{W}}_m}\sum_{j=1}^{n_I}{\sum_{l=1}^{n_s}{\tilde{b}_{mjl}^{k}\cdot \left( \mathbf{\tilde{W}}_{m}^{'}\mathbf{f}_{mjl}^{k} \right)}},
\end{equation}
where $\mathbf{\tilde{W}}_m$ and $\mathbf{\tilde{W}}_{m}^{'}$ are learnable weights. As in QGSLF, we combine original and image-enhanced grid features as
\begin{equation}\label{equation27}
\mathbf{F}_{gfuse}^{k}=\text{FFN}\left( \text{Concat}\left( \mathbf{F}_{g}^{k},\ \mathbf{\hat{F}}_{g}^{k} \right) \right) .
\end{equation}

Repeating the aforementioned operations for all grids in  $
\mathbf{B}_i$, we obtain the semantic-enhanced proposal feature $
\mathbf{F}_{B}^{i}$. Therefore, the proposal-level features for all proposals processed by the QGPLF block can be denoted as
\begin{equation}\label{equation28}
\mathbf{F}_{PLP}=\left[ \mathbf{F}_{B}^{1},\cdots,\mathbf{F}_{B}^{N_{proposal}} \right] ,
\end{equation}
which incorporate locally geometric information derived from radar points and semantic information derived from images.

To further refine the proposal features, $\mathbf{F}_{PLP}$ is fused with pooled scene-level features $\mathbf{F}_{SLP}$ from the HSFP module via the attention operations. While $\mathbf{F}_{PLP}$ derives from proposal-image fusion, $\mathbf{F}_{SLP}$ is pooled from the fused scene-level features, capturing scene context. Their complementary integration enhances the refined proposal features with both instance-specific details and scene contextual information. Concretely, as shown in Fig. \ref{figure7}, the features $\mathbf{F}_{PLP}$ and $\mathbf{F}_{SLP}$ are concatenated and processed through a self-attention layer:
\begin{equation}\label{equation29}
\mathbf{X}_P=\text{MSA}\left( \text{Concat}\left( \mathbf{F}_{PLP},\mathbf{F}_{SLP} \right) \right) ,
\end{equation}
where MSA denotes Multi-Head Self-Attention. The refined proposal feature $\mathbf{X}_P$ is fed into the detection head.

\subsection{Detection Head}
The detection head takes $\mathbf{X}_P$ in (\ref{equation29}) from PLFE as input for the proposal refinement network. Concretely, the refinement network adopts a shared two-layer MLP and has two parallel branches for bounding box regression and confidence prediction, respectively. Following the approaches in \cite{shi2020pv, deng2021voxel}, the box regression branch predicts the residue from 3D region proposals to the ground truth (GT) boxes, while the confidence branch predicts the confidence score.

\subsection{Loss Function}
In this paper, the weights of the image encoder are pretrained and remain frozen during training. We only train the radar branch with the region proposal loss $\mathcal{L}_{\text{RPN}}$ and the proposal refinement losses, which include the confidence prediction loss $
\mathcal{L}_{\text{conf}}$ and the box regression loss $
\mathcal{L}_{\text{reg}}$, consistent with \cite{deng2021voxel, shi2020pv}. The overall training loss is computed as the sum of these losses, i.e.,
\begin{equation}\label{equation30}
\mathcal{L}=\mathcal{L}_{\text{RPN}}+\mathcal{L}_{\text{conf}}+\mathcal{L}_{\text{reg}}.
\end{equation}

\section{Experiments}
\subsection{Datasets and Evaluation Metrics}
\subsubsection{Datasets}
Following prior studies \cite{zheng2023rcfusion, gu2025hgsfusion, liu2025mssf}, we perform extensive experiments on two publicly available 4D radar datasets, i.e., the VoD dataset and the TJ4DRadSet dataset. Both datasets serve autonomous driving research, specifically advancing 4D radar perception tasks.

The VoD dataset was collected in Delft, Netherlands, covering urban scenarios such as campuses, suburbs, and old towns. It provides synchronized multi-modal data, including 4D radar, LiDAR, camera data with 3D annotations. Furthermore, it also contains 4D radar point clouds accumulated from multiple scans with ego-motion compensation, enhancing point cloud density. The VoD dataset comprises 8682 frames, officially split into 5139 training frames, 1296 validation frames, and 2247 testing frames. As the official evaluation system remains unavailable, we follow prior works \cite{zheng2023rcfusion, gu2025hgsfusion, liu2025mssf} in conducting comparison and ablation experiments on the validation set. Specifically, we evaluate the detection results of car, pedestrian, and cyclist categories using five-scan accumulated 4D radar data fused with camera inputs.

The TJ4DRadSet dataset, collected in Suzhou, China, encompasses more complex scenarios including urban roads, elevated highways, industrial zones, and challenging conditions such as nighttime, glare, and tunnel shadows. Different from the VoD dataset, this dataset currently offers only 4D radar and camera synchronization. It contains 7757 frames, officially split into 5717 training and 2040 testing frames. In our experiments, we evaluate detection performances for the four object categories: car, pedestrian, cyclist, and truck.

\subsubsection{Evaluation Metrics}
For the VoD dataset, two evaluation metrics are employed: Average Precision (AP) in the Entire Annotated Area (EAA), which considers all annotated objects regardless of their ranges, and AP in the Driving Corridor Area (DCA), which only evaluates objects within the region defined as 
$\text{DCA}=\{\left( x,y,z \right) |-4\text{m}<x<4\text{m},\ z<25\text{m}\}$. The mean AP (mAP) is calculated by averaging the AP results over multiple categories. Following the official setting, the intersection-over-union (IoU) thresholds for calculating AP are set to 0.25 for pedestrians and cyclists, and 0.5 for cars.

For the TJ4DRadSet dataset, evaluation metrics comprise AP in 3D space ($\text{AP}_{3\text{D}}$) and AP in BEV space ($
\text{AP}_{\text{BEV}}$), both evaluated for objects within a 70-meter range of the ego-vehicle. Similar to those used in the VoD dataset, IoU thresholds are set to 0.25 for pedestrians and cyclists, and 0.5 for cars. Additionally, an IoU threshold of 0.5 is applied to the truck category.

\subsection{Implementation Details}
We follow the official configurations for each dataset. For the VoD dataset, the radar point cloud range is set to $
\{\left( x,y,z \right) |0\text{m}<x<51.2\text{m},-25.6\text{m}<y<25.6\text{m},-3\text{m}<z<2\text{m}\}$. Voxel sizes are set to 0.05m, 0.05m, and 0.125m along the X-, Y-, and Z-axis, respectively. Anchor box dimensions for the car, pedestrian, and cyclist categories are (3.9m, 1.6m, 1.56m), (0.8m, 0.6m, 1.73m), and (1.76m, 0.6m, 1.73m), respectively. We use the 5-scan accumulated radar point cloud data as the input. Each point can be represented by a 7-dimensional feature vector:    
\begin{equation}\label{equation31}
\mathbf{f}_{raw}^{\text{VoD}}=\left[ x,y,z,RCS,v_r,v_{rc},t \right] ,
\end{equation}
where $RCS$ denotes the reflectivity of the detection, $v_r$ is the relative radial Doppler velocity, $v_{rc}$ represents the absolute radial Doppler velocity, and $t$ is the temporal scan identifier.

Similarly, for the TJ4DRadSet dataset, the radar point cloud range is set to $
\{\left( x,y,z \right) |0\text{m}<x<69.12\text{m},-39.68\text{m}<y<39.68\text{m},-4\text{m}<z<2\text{m}\}
$. Voxel sizes are set to 0.08m, 0.08m, and 0.125m along the X-, Y-, and Z-axis, respectively. Anchor box dimensions for the car, pedestrian, cyclist, and truck categories are (4.56m, 1.84m, 1.70m), (0.8m, 0.6m, 1.69m), (1.77m, 0.78m, 1.60m), and (10.76m, 2.66m, 3.47m), respectively. We utilize the original 8-dimensional features of radar points as the input, which can be denoted as
\begin{equation}\label{equation32}
\mathbf{f}_{\text{raw}}^{\text{TJ4D}}=\left[ x,y,z,v_r,Range,Power,Alpha,Beta \right] ,
\end{equation}
where $v_r$ is the relative radial Doppler velocity, $Range$ is the detection range to radar center, $Power$ in dB scale is the signal to noise ratio of the detection, and $Alpha$ and $Beta$ are horizontal and vertical angles of the detection, respectively.

In both datasets, ResNet-101 \cite{he2016deep} is used as the image backbone. Its weights, pretrained on DeepLabV3, remain frozen during training. Within the ERPE module of our model, Mask2former \cite{cheng2022masked} is employed as the segmentation network for generating 2D instances, and its weights remain frozen during training. The radius $r$ is set to 51, the standard deviations $
\sigma _1$ and $\sigma _2$ are set to 7, and the fixed number $\tau$ of sampled points is 50 following \cite{gu2025hgsfusion}. In the HSFP module, we integrate the downsampled 4× and 8× voxel feature volumes $\mathcal{X}_3$ and $\mathcal{X}_4$ with multi-scale image features from 5 levels. The deformable attention mechanism in this module uses 4 sampling points and 4 attention heads. In the PLFE module, each proposal is partitioned into a grid of size $6\times 6\times 6$. Similarly, the deformable attention operation employs 4 sampling points and 4 attention heads, and also incorporates multi-scale image features from 5 levels.

We implement our model based on the OpenPCDet framework. The model is trained on two NVIDIA A800 GPUs using the AdamW optimizer. We employ a one-cycle learning rate policy with an initial learning rate of 0.001. For the input 4D radar point clouds, we apply data augmentation including random flipping, scaling, and rotation. No augmentation is applied to the input images.

\begin{table*}[pos=t]\rmfamily
\begin{center}
\caption{Performance comparison on the validation set of the VoD dataset.}
\label{Table1}
\setlength{\tabcolsep}{1.1mm}
\renewcommand{\arraystretch}{1.2}
\begin{tabular}{cll|ccccccccl}
\toprule
\multicolumn{3}{c}{\multirow{2}{*}{Method}}        & \multirow{2}{*}{Modality} & \multicolumn{4}{c}{AP in the Entire Annotated Area (\%)}                                                                        & \multicolumn{4}{c}{AP in the Driving Corridor (\%)}                                                                                       \\ \cmidrule(r){5-8}
\cmidrule(r){9-12}
\multicolumn{3}{c}{}                               &                           & \multicolumn{1}{c}{Car}            & \multicolumn{1}{c}{Pedestrian}     & \multicolumn{1}{c}{Cyclist}        & mAP            & \multicolumn{1}{c}{Car}            & \multicolumn{1}{c}{Pedestrian}     & \multicolumn{1}{c}{Cyclist}        & \multicolumn{1}{c}{mAP} \\ \hline
\multicolumn{3}{c}{PointPillars (CVPR2019) \cite{lang2019pointpillars}}        & R                         & \multicolumn{1}{c}{37.06}          & \multicolumn{1}{c}{35.04}          & \multicolumn{1}{c}{63.44}          & 45.18          & \multicolumn{1}{c}{70.15}          & \multicolumn{1}{c}{47.22}          & \multicolumn{1}{c}{85.07}          & 67.48                    \\ 
\multicolumn{3}{c}{RadarPillarNet (IEEE TIM 2023) \cite{zheng2023rcfusion}} & R                         & \multicolumn{1}{c}{39.30}          & \multicolumn{1}{c}{35.10}          & \multicolumn{1}{c}{63.63}          & 46.01          & \multicolumn{1}{c}{71.65}          & \multicolumn{1}{c}{42.80}          & \multicolumn{1}{c}{83.14}          & 65.86                    \\ 
\multicolumn{3}{c}{VoxelNeXt (CVPR 2023) \cite{chen2023voxelnext}}          & R                         & \multicolumn{1}{c}{36.98}          & \multicolumn{1}{c}{42.37}          & \multicolumn{1}{c}{68.15}          & 49.17          & \multicolumn{1}{c}{70.95}          & \multicolumn{1}{c}{51.14}          & \multicolumn{1}{c}{85.67}          & 70.04                    \\ 
\multicolumn{3}{c}{SMURF (IEEE TIV 2023) \cite{liu2023smurf}}          & R                         & \multicolumn{1}{c}{42.31}          & \multicolumn{1}{c}{39.09}          & \multicolumn{1}{c}{71.50}          & 50.97          & \multicolumn{1}{c}{71.74}          & \multicolumn{1}{c}{50.54}          & \multicolumn{1}{c}{86.87}          & 69.72                    \\ 
\multicolumn{3}{c}{MVFAN (ICONIP 2023) \cite{yan2023mvfan}}            & R                         & \multicolumn{1}{c}{34.05}          & \multicolumn{1}{c}{27.27}          & \multicolumn{1}{c}{57.14}          & 39.42          & \multicolumn{1}{c}{69.81}          & \multicolumn{1}{c}{38.65}          & \multicolumn{1}{c}{84.87}          & 64.38                    \\ 
\multicolumn{3}{c}{MUFASA (ICANN 2024) \cite{peng2024mufasa}}            & R                         & \multicolumn{1}{c}{43.10}          & \multicolumn{1}{c}{38.97}          & \multicolumn{1}{c}{68.65}          & 50.24          & \multicolumn{1}{c}{72.50}          & \multicolumn{1}{c}{50.28}          & \multicolumn{1}{c}{88.51}          & 70.43                    \\ 
\multicolumn{3}{c}{RadarPillars (IEEE ITSC 2024) \cite{musiat2024radarpillars}}  & R                         & \multicolumn{1}{c}{41.10}          & \multicolumn{1}{c}{38.60}          & \multicolumn{1}{c}{72.60}          & 50.70          & \multicolumn{1}{c}{71.10}          & \multicolumn{1}{c}{52.30}          & \multicolumn{1}{c}{87.90}          & 70.50                    \\ \hline
\multicolumn{3}{c}{FUTR3D (CVPR 2023) \cite{chen2023futr3d}}             & R+C                       & \multicolumn{1}{c}{46.01}          & \multicolumn{1}{c}{35.11}          & \multicolumn{1}{c}{65.98}          & 49.03          & \multicolumn{1}{c}{78.66}          & \multicolumn{1}{c}{43.10}          & \multicolumn{1}{c}{86.19}          & 69.32                    \\ 
\multicolumn{3}{c}{BEVFusion (ICRA 2023) \cite{liu2023bevfusion}}          & R+C                       & \multicolumn{1}{c}{37.85}          & \multicolumn{1}{c}{40.96}          & \multicolumn{1}{c}{68.95}          & 49.25          & \multicolumn{1}{c}{70.21}          & \multicolumn{1}{c}{45.86}          & \multicolumn{1}{c}{89.48}          & 68.52                    \\ 
\multicolumn{3}{c}{RCFusion (IEEE TIM 2023) \cite{zheng2023rcfusion}}       & R+C                       & \multicolumn{1}{c}{41.70}          & \multicolumn{1}{c}{38.95}          & \multicolumn{1}{c}{68.31}          & 49.65          & \multicolumn{1}{c}{71.87}          & \multicolumn{1}{c}{47.50}          & \multicolumn{1}{c}{88.33}          & 69.23                    \\ 
\multicolumn{3}{c}{TL-4DRCF (IEEE Sens.J 2024) \cite{zhang2024tl}}    & R+C                       & \multicolumn{1}{c}{43.71}          & \multicolumn{1}{c}{40.11}          & \multicolumn{1}{c}{64.22}          & 49.35          & \multicolumn{1}{c}{79.49}          & \multicolumn{1}{c}{53.76}          & \multicolumn{1}{c}{76.50}          & 69.92                    \\ 
\multicolumn{3}{c}{RCBEVDet (CVPR 2024) \cite{lin2024rcbevdet}}           & R+C                       & \multicolumn{1}{c}{40.63}          & \multicolumn{1}{c}{38.86}          & \multicolumn{1}{c}{70.68}          & 49.99          & \multicolumn{1}{c}{72.48}          & \multicolumn{1}{c}{49.89}          & \multicolumn{1}{c}{87.01}          & 69.80                    \\ 
\multicolumn{3}{c}{SGDet3D (IEEE RAL 2024) \cite{bai2024sgdet3d}}        & R+C                       & \multicolumn{1}{c}{53.16}          & \multicolumn{1}{c}{49.98}          & \multicolumn{1}{c}{76.11}          & 59.75          & \multicolumn{1}{c}{81.13}          & \multicolumn{1}{c}{60.93}          & \multicolumn{1}{c}{90.22}          & 77.42                    \\ 
\multicolumn{3}{c}{DSFusion (EAAI 2025) \cite{bi2025dual}}           & R+C                       & \multicolumn{1}{c}{45.75}          & \multicolumn{1}{c}{50.71}          & \multicolumn{1}{c}{72.64}          & 56.37          & \multicolumn{1}{c}{78.21}          & \multicolumn{1}{c}{61.22}          & \multicolumn{1}{c}{87.71}          & 73.16                    \\ 
\multicolumn{3}{c}{MSSF (IEEE TITS 2025) \cite{liu2025mssf}}          & R+C                       & \multicolumn{1}{c}{52.53}          & \multicolumn{1}{c}{51.31}          & \multicolumn{1}{c}{75.77}          & 59.96          & \multicolumn{1}{c}{89.08}          & \multicolumn{1}{c}{\textbf{66.78}} & \multicolumn{1}{c}{88.10}          & 81.32                    \\ 
\multicolumn{3}{c}{HGSFusion (AAAI 2025) \cite{gu2025hgsfusion}}          & R+C                       & \multicolumn{1}{c}{51.67}          & \multicolumn{1}{c}{\textbf{52.64}} & \multicolumn{1}{c}{72.58}          & 58.96          & \multicolumn{1}{c}{88.28}          & \multicolumn{1}{c}{62.61}          & \multicolumn{1}{c}{87.49}          & 79.46                    \\ 
\multicolumn{3}{c}{\textbf{MLF-4DRCNet (ours)}}    & R+C                       & \multicolumn{1}{c}{\textbf{53.29}} & \multicolumn{1}{c}{51.36}          & \multicolumn{1}{c}{\textbf{76.17}} & \textbf{60.28} & \multicolumn{1}{c}{\textbf{90.20}} & \multicolumn{1}{c}{61.83}          & \multicolumn{1}{c}{\textbf{95.66}} & \textbf{82.57}           \\ \bottomrule
\end{tabular}
\begin{tablenotes} 
\item The symbols R and C denote 4D radar and camera, respectively. Best values are shown in bold.
\end{tablenotes} 
\end{center}
\end{table*}

\subsection{Comparisons with SOTA}
To comprehensively evaluate the performance of the proposed MLF-4DRCNet, we compare it against SOTA methods on the VoD and TJ4DRadSet datasets. 

\subsubsection{Results on VoD}
Table \ref{Table1} presents the performance of different methods on the validation set of the VoD dataset. The observed performance gap between EAA and DCA clearly demonstrates that 4D radar achieves better detection performance for nearby objects, as these objects reflect denser radar points at close range. Furthermore, the overall experimental results reveal that our MLF-4DRCNet achieves SOTA performance, with an overall mAP of 60.28\% in the EAA and a remarkable 82.57\% in the DCA. Compared to radar-only methods such as SMURF\cite{liu2023smurf} and RadarPillars \cite{musiat2024radarpillars}, our model achieves a notable increase in mAP by at least 9.31\% in the EAA and 12.07\% in the DCA. These results confirm that MLF-4DRCNet outperforms all 4D radar-only methods in 3D object detection by leveraging complementary information from camera images. When compared to recent radar-camera fusion methods, such as HGSFusion \cite{gu2025hgsfusion} and MSSF \cite{liu2025mssf}, our model still achieves superior performance across nearly all evaluation metrics, particularly in detecting cars and cyclists. MLF-4DRCNet enhances the mAP value by at least 0.32\% and 1.25\% in the EAA and DCA, respectively. The outstanding performance can be attributed to the effective multi-level fusion of multi-modal information. It is noteworthy that our method does not improve significantly on the pedestrian category, which we attribute to the fact that the reflections of non-metallic objects to millimeter waves are much weaker than those of metallic objects. During the experiments, we also find that pedestrians and cyclists in the VoD dataset often appear close together and are challenging to distinguish under occlusion, which adversely affects the quality of virtual points and consequently degrades detection accuracy.

\begin{figure*}[pos=t]
\centering
\includegraphics[scale=0.27]{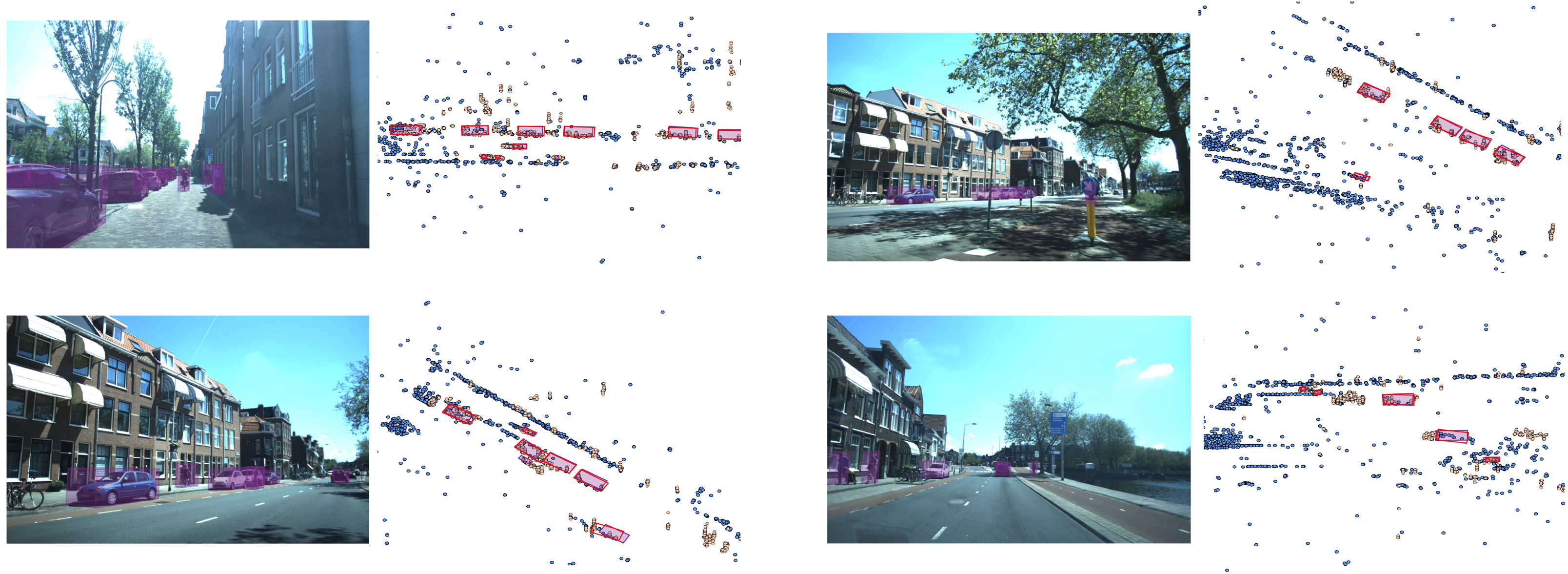}
\caption{Visualization results on the VoD dataset under the BEV perspective. Detection results from the proposed MLF-4DRCNet are indicated by red boxes, while ground truth annotations are denoted by purple boxes. Raw radar points are displayed in blue, and the generated virtual points are shown in yellow.}
\label{figure9}
\end{figure*}

Visualization results of the VoD dataset are presented in Fig. \ref{figure9}, along with the generated virtual points. One can see that these virtual points increase the density of radar points in the object regions, thus improving detection probability.

\begin{table*}[pos=ht]\rmfamily
\begin{center}
\caption{Performance comparison on the test set of the TJ4DRadSet dataset.}
\label{Table2}
\setlength{\tabcolsep}{0.7mm}
\renewcommand{\arraystretch}{1.2}
\begin{tabular}{cll|ccccccccccc}
\toprule
\multicolumn{3}{c}{\multirow{2}{*}{Method}}        & \multirow{2}{*}{Modality} & \multicolumn{5}{c}{$
\text{AP}_{3\text{D}}$ (\%)}                                                                                                                                       & \multicolumn{5}{c}{$
\text{AP}_{\text{BEV}}$ (\%)}                                                                                                                                        \\ \cmidrule(r){5-9}
\cmidrule(r){10-14}
\multicolumn{3}{c}{}                               &                           & \multicolumn{1}{c}{Car}            & \multicolumn{1}{c}{Pedestrian}     & \multicolumn{1}{c}{Cyclist}        & \multicolumn{1}{c}{Truck}          & mAP            & \multicolumn{1}{c}{Car}            & \multicolumn{1}{c}{Pedestrian}     & \multicolumn{1}{c}{Cyclist}        & \multicolumn{1}{c}{Truck}          & mAP            \\ \hline
\multicolumn{3}{c}{PointPillars (CVPR 2019) \cite{lang2019pointpillars}}       & R                         & \multicolumn{1}{c}{19.78}          & \multicolumn{1}{c}{29.79}          & \multicolumn{1}{c}{51.83}          & \multicolumn{1}{c}{13.67}          & 28.76          & \multicolumn{1}{c}{39.42}          & \multicolumn{1}{c}{32.56}          & \multicolumn{1}{c}{59.45}          & \multicolumn{1}{c}{20.93}          & 38.09          \\ 
\multicolumn{3}{c}{RadarPillarNet (IEEE TIM 2023) \cite{zheng2023rcfusion}} & R                         & \multicolumn{1}{c}{28.45}          & \multicolumn{1}{c}{26.24}          & \multicolumn{1}{c}{51.57}          & \multicolumn{1}{c}{15.20}          & 30.37          & \multicolumn{1}{c}{45.72}          & \multicolumn{1}{c}{29.19}          & \multicolumn{1}{c}{56.89}          & \multicolumn{1}{c}{25.17}          & 39.24          \\ 
\multicolumn{3}{c}{VoxelNeXt (CVPR 2023) \cite{chen2023voxelnext}}          & R                         & \multicolumn{1}{c}{13.27}          & \multicolumn{1}{c}{33.54}          & \multicolumn{1}{c}{52.59}          & \multicolumn{1}{c}{8.32}           & 26.93          & \multicolumn{1}{c}{23.17}          & \multicolumn{1}{c}{35.83}          & \multicolumn{1}{c}{57.11}          & \multicolumn{1}{c}{12.12}          & 32.06          \\ 
\multicolumn{3}{c}{SMURF (IEEE TIV 2023) \cite{liu2023smurf}}          & R                         & \multicolumn{1}{c}{28.47}          & \multicolumn{1}{c}{26.22}          & \multicolumn{1}{c}{54.61}          & \multicolumn{1}{c}{22.64}          & 32.99          & \multicolumn{1}{c}{43.13}          & \multicolumn{1}{c}{29.19}          & \multicolumn{1}{c}{58.81}          & \multicolumn{1}{c}{32.80}          & 40.98          \\ \hline
\multicolumn{3}{c}{RCFusion (IEEE TIM 2023) \cite{zheng2023rcfusion}}       & R+C                       & \multicolumn{1}{c}{29.72}          & \multicolumn{1}{c}{27.17}          & \multicolumn{1}{c}{54.93}          & \multicolumn{1}{c}{23.56}          & 33.85          & \multicolumn{1}{c}{40.89}          & \multicolumn{1}{c}{30.95}          & \multicolumn{1}{c}{58.30}          & \multicolumn{1}{c}{28.92}          & 39.76          \\ 
\multicolumn{3}{c}{FUTR3D (CVPR 2023) \cite{chen2023futr3d}}             & R+C                       & \multicolumn{1}{c}{-}              & \multicolumn{1}{c}{-}              & \multicolumn{1}{c}{-}              & \multicolumn{1}{c}{-}              & 32.42          & \multicolumn{1}{c}{-}              & \multicolumn{1}{c}{-}              & \multicolumn{1}{c}{-}              & \multicolumn{1}{c}{-}              & 37.51          \\ 
\multicolumn{3}{c}{BEVFusion (ICRA 2023) \cite{liu2023bevfusion}}          & R+C                       & \multicolumn{1}{c}{-}              & \multicolumn{1}{c}{-}              & \multicolumn{1}{c}{-}              & \multicolumn{1}{c}{-}              & 32.71          & \multicolumn{1}{c}{-}              & \multicolumn{1}{c}{-}              & \multicolumn{1}{c}{-}              & \multicolumn{1}{c}{-}              & 41.12          \\ 
\multicolumn{3}{c}{DSFusion (EAAI 2025) \cite{bi2025dual}}           & R+C                       & \multicolumn{1}{c}{37.76}          & \multicolumn{1}{c}{\textbf{34.09}}          & \multicolumn{1}{c}{49.88}          & \multicolumn{1}{c}{26.87}          & 37.15          & \multicolumn{1}{c}{49.86}          & \multicolumn{1}{c}{34.17}          & \multicolumn{1}{c}{52.49}          & \multicolumn{1}{c}{33.14}          & 42.41          \\ 
\multicolumn{3}{c}{MSSF (IEEE TITS 2025) \cite{liu2025mssf}}          & R+C                       & \multicolumn{1}{c}{45.18}          & \multicolumn{1}{c}{33.61}          & \multicolumn{1}{c}{55.88}          & \multicolumn{1}{c}{17.20}          & 37.97          & \multicolumn{1}{c}{56.25}          & \multicolumn{1}{c}{\textbf{36.53}} & \multicolumn{1}{c}{\textbf{58.70}} & \multicolumn{1}{c}{20.97}          & 43.11          \\ 
\multicolumn{3}{c}{HGSFusion (AAAI 2025) \cite{gu2025hgsfusion}}          & R+C                       & \multicolumn{1}{c}{-}              & \multicolumn{1}{c}{-}              & \multicolumn{1}{c}{-}              & \multicolumn{1}{c}{-}              & 37.21          & \multicolumn{1}{c}{-}              & \multicolumn{1}{c}{-}              & \multicolumn{1}{c}{-}              & \multicolumn{1}{c}{-}              & 43.23          \\ 
\multicolumn{3}{c}{\textbf{MLF-4DRCNet (ours)}}    & R+C                       & \multicolumn{1}{c}{\textbf{49.71}} & \multicolumn{1}{c}{33.93} & \multicolumn{1}{c}{\textbf{56.51}} & \multicolumn{1}{c}{\textbf{30.25}} & \textbf{42.60} & \multicolumn{1}{c}{\textbf{57.79}} & \multicolumn{1}{c}{35.94}          & \multicolumn{1}{c}{57.35}          & \multicolumn{1}{c}{\textbf{38.71}} & \textbf{47.45} \\ \bottomrule
\end{tabular}
\begin{tablenotes} 
\item The symbols R and C denote 4D radar and camera, respectively. Best values are shown in bold.
\end{tablenotes} 
\end{center}
\end{table*}

\subsubsection{Results on TJ4DRadSet}
\begin{figure*}[pos=t]
\centering
\includegraphics[scale=0.27]{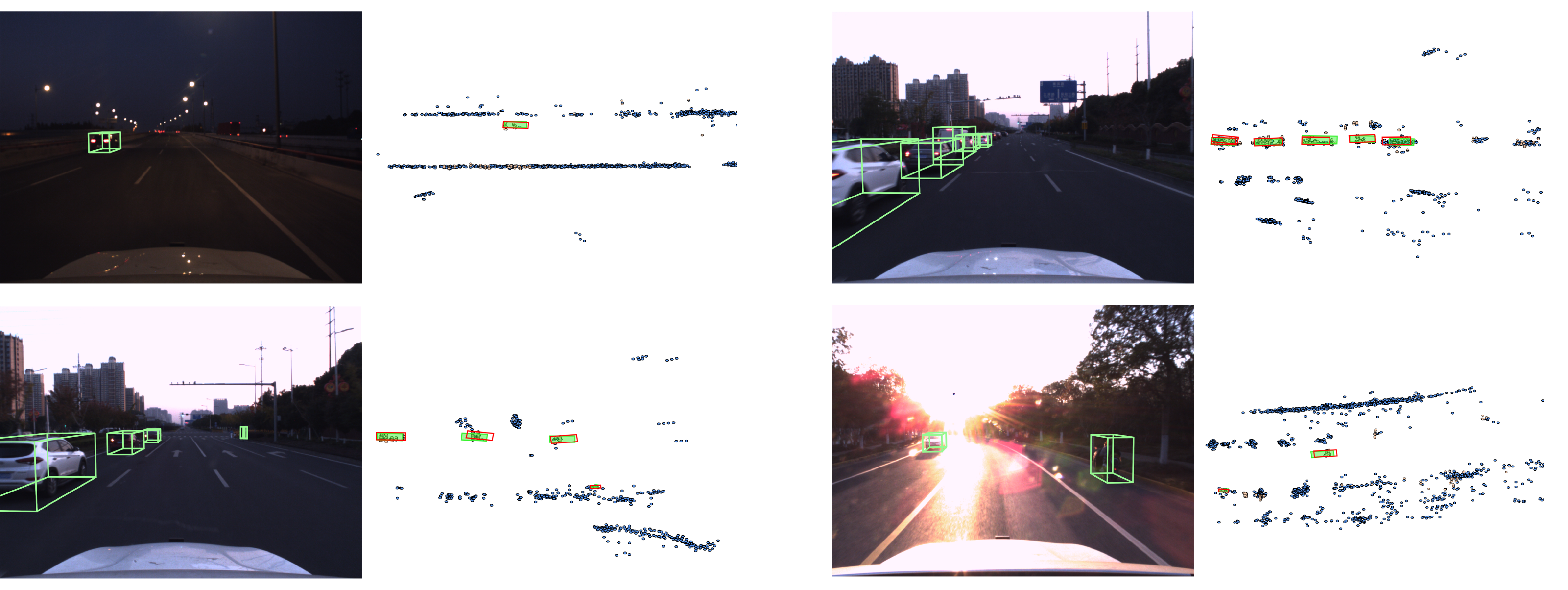}
\caption{Visualization results on the TJ4DRadSet dataset under the BEV perspective. Detection results from the proposed MLF-4DRCNet are indicated by red boxes, while ground truth annotations are denoted by green boxes. Raw radar points are displayed in blue, and the generated virtual points are shown in yellow.}
\label{figure10}
\end{figure*}

We further validate the generalization ability of MLF-4DRCNet on the TJ4DRadSet test set. Compared with VoD, TJ4DRadSet presents a greater challenge for covering more complex scenarios, such as low-light night-time highways and highly illuminated urban roads, where image quality degrades significantly. Moreover, TJ4DRadset introduces an additional truck category, in which object sizes vary significantly, further increasing the detection difficulty. TJ4DRadSet does not utilize multi-frame accumulation, resulting in sparser point clouds than those in VoD. Despite these challenges, our MLF-4DRCNet also achieves SOTA performance (see Table \ref{Table2}), attaining the highest overall 3D mAP of 42.60\% and BEV mAP of 47.45\%, surpassing all other radar-only and fusion-based methods. Notably, for the challenging truck category, our method improves $\text{AP}_{3\text{D}}$ and $
\text{AP}_{\text{BEV}}$ by at least 3.38\% and 5.91\%, respectively. The reason is that our method can significantly densify radar point clouds of trucks and effectively integrate multi-level features from both radar and camera data. Overall, these results in Table \ref{Table2} validate that our MLF-4DRCNet can effectively exploit multi-modal information to attain superior performance even in challenging and complex environments. Fig. \ref{figure10} shows the visualization results for the TJ4DRadSet dataset, in a style similar to that used for the VoD dataset.

\begin{table}[pos=ht]\rmfamily
\begin{center}
\caption{Results of ablation experiments for ERPE.}
\setlength{\tabcolsep}{3mm}
\renewcommand{\arraystretch}{1.2}
\label{Table3}
\begin{tabular}{cclcc}
\toprule
Virtual Points & \multicolumn{2}{c}{Voxel Encoding Strategy}            & $\text{mAP}_{\text{EAA}}$ & $\text{mAP}_{\text{DCA}}$ \\ \hline
               & \multicolumn{2}{c}{Mean VFE}                     & 57.47      & 77.64      \\ 
\checkmark              & \multicolumn{2}{c}{Mean VFE}                     & 58.87      & 79.72      \\ 
\checkmark              & \multicolumn{2}{c}{Separate Encoding + Mean VFE} & 59.66      & 81.33      \\ 
\checkmark             & \multicolumn{2}{c}{TA VFE (ours)}                & \textbf{60.28}      & \textbf{82.57}      \\ \bottomrule
\end{tabular}
\end{center}
\end{table}

\subsection{Ablation Studies}
To validate the efficacy of each component of MLF-4DRCNet, we conduct comprehensive ablation studies on the VoD validation set.
\subsubsection{Effect of ERPE}
The ERPE module is designed to densify the raw point cloud using virtual points and subsequently perform voxel encoding. Therefore, we evaluate the effectiveness of ERPE based on whether virtual points are employed and which voxel encoding strategy is used. As shown in the Table \ref{Table3}, introducing virtual points in raw radar points consistently improves performance in both EAA and DCA. This is attributable to the semantic information from images introduced by the virtual points, enhancing the quality of the radar point clouds. Additionally, we employ different voxel encoding strategies to evaluate the effectiveness of the proposed TA-VFE, including the commonly used Mean-VFE and its enhanced version by incorporating separate encoding \cite{gu2025hgsfusion}. It can be observed that the best results are obtained with the proposed TA-VFE, which reaches 60.28\% mAP in EAA and 82.57\% in DCA. This validates that TA-VFE can more effectively utilize the beneficial information introduced by virtual points than the normal voxel encoding strategy. It employs three different types of attention weighting to enhance critical information in the hybrid point clouds while suppressing the inevitable noise introduced by the virtual points, thus yielding more robust voxel features.

\subsubsection{Effect of HSFP}
The HSFP module integrates multi-scale voxel features and image features at the scene level, generating semantically enhanced fused features. Therefore, we examine the effectiveness of HSFP by exploring the contribution of voxel features with different downsampling rates. As shown in Table \ref{Table4}, $\mathcal{X}_1$, $\mathcal{X}_2$, $\mathcal{X}_3$, and $\mathcal{X}_4$ correspond to voxel features with 1×, 2×, 4×, and 8× downsampling rates respectively. The results demonstrate that HSFP significantly improves detection performance in both EAA and DCA. In particular, incorporating all four features yields the highest mAP of 60.72\% in the EAA. In contrast, although using only $\mathcal{X}_3$ and $\mathcal{X}_4$ leads to a slight decrease in EAA performance, it achieves the highest mAP of 82.57\% in the DCA. The reason is that the detection range of DCA is closer to the ego vehicle than EAA. Features with lower downsampling rates retain more spatial details, which are beneficial for large-area detection but may introduce unnecessary information or noise that hinders performance in closer-range scenarios. On the other hand, highly downsampled features provide richer semantic contexts essential for object recognition. The hierarchical fusion strategy in HSFP effectively integrates multi-scale representations, optimizing overall detection performance. Given that DCA is considered more critical than EAA, we use $\mathcal{X}_3$ and $\mathcal{X}_4$ for HSFP in our model.

\begin{table}[pos=t]\rmfamily
\begin{center}
\caption{Results of ablation experiments for HSFP.}
\setlength{\tabcolsep}{3.2mm}
\renewcommand{\arraystretch}{1.05}
\label{Table4}
\begin{tabular}{cccccc}
\toprule
\multicolumn{4}{c}{Voxel Features Used for HSFP}                         & \multirow{2}{*}{$\text{mAP}_{\text{EAA}}$} & \multirow{2}{*}{$\text{mAP}_{\text{DCA}}$} \\ \cline{1-4}
\multicolumn{1}{c}{$\mathcal{X}_1$} & \multicolumn{1}{c}{$\mathcal{X}_2$} & \multicolumn{1}{c}{$\mathcal{X}_3$} & \multicolumn{1}{c}{$\mathcal{X}_4$} &                      &                      \\ \hline
\multicolumn{1}{c}{}   & \multicolumn{1}{c}{}   & \multicolumn{1}{c}{}   &    & 58.08                & 76.34                \\ 
\multicolumn{1}{c}{\checkmark}  & \multicolumn{1}{c}{\checkmark}  & \multicolumn{1}{c}{\checkmark}  & \checkmark  & \textbf{60.72}                & 79.96                \\ 
\multicolumn{1}{c}{}   & \multicolumn{1}{c}{\checkmark}  & \multicolumn{1}{c}{\checkmark}  & \checkmark  & 60.35       & 80.12                \\ 
\multicolumn{1}{c}{}   & \multicolumn{1}{c}{}   & \multicolumn{1}{c}{\checkmark}  & \checkmark  & 60.28                & \textbf{82.57}       \\ 
\multicolumn{1}{c}{}   & \multicolumn{1}{c}{}   & \multicolumn{1}{c}{}  & \checkmark  & 59.83                & 82.03                \\ \bottomrule
\end{tabular}
\end{center}
\end{table}

\begin{table}\rmfamily
\begin{center}
\caption{Results of ablation experiments for PLFE.}
\label{Table5}
\renewcommand{\arraystretch}{1.2}
\setlength{\tabcolsep}{1.2mm}
\begin{tabular}{ccccccccc}
\toprule
\multirow{2}{*}{Module} & \multicolumn{4}{c}{AP in the EAA (\%)}                                                                                          & \multicolumn{4}{c}{AP in the DCA (\%)}                                                                                          \\ \cmidrule(r){2-5}
\cmidrule(r){6-9}
                        & \multicolumn{1}{c}{Car}            & \multicolumn{1}{c}{Pedestrian}     & \multicolumn{1}{c}{Cyclist}        & mAP            & \multicolumn{1}{c}{Car}            & \multicolumn{1}{c}{Pedestrian}     & \multicolumn{1}{c}{Cyclist}        & mAP            \\ \hline
No PLFE                 & \multicolumn{1}{c}{53.08}          & \multicolumn{1}{c}{50.79}          & \multicolumn{1}{c}{74.43}          & 58.86          & \multicolumn{1}{c}{89.90}          & \multicolumn{1}{c}{61.12}          & \multicolumn{1}{c}{92.96}          & 81.32          \\ 
PLFE                    & \multicolumn{1}{c}{\textbf{53.29}} & \multicolumn{1}{c}{\textbf{51.36}} & \multicolumn{1}{c}{\textbf{76.17}} & \textbf{60.28} & \multicolumn{1}{c}{\textbf{90.20}} & \multicolumn{1}{c}{\textbf{61.83}} & \multicolumn{1}{c}{\textbf{95.66}} & \textbf{82.57} \\ \bottomrule
\end{tabular}
\end{center}
\end{table}

\begin{table*}\rmfamily
\begin{center}
\caption{Performance comparison on the validation set of the VoD dataset.}
\label{Table6}
\renewcommand{\arraystretch}{1.2}
\setlength{\tabcolsep}{1mm}
\begin{tabular}{cllccccccccl}
\toprule
\multicolumn{3}{c}{\multirow{2}{*}{Method}}        & \multirow{2}{*}{Modality} & \multicolumn{4}{c}{AP in the EAA (\%)}                                                                        & \multicolumn{4}{c}{AP in the DCA (\%)}                                                                                       \\ \cmidrule(r){5-8}
\cmidrule(r){9-12}
\multicolumn{3}{c}{}                               &                           & \multicolumn{1}{c}{Car}            & \multicolumn{1}{c}{Pedestrian}     & \multicolumn{1}{c}{Cyclist}        & mAP            & \multicolumn{1}{c}{Car}            & \multicolumn{1}{c}{Pedestrian}     & \multicolumn{1}{c}{Cyclist}        & \multicolumn{1}{c}{mAP} \\ \hline
\multicolumn{3}{c}{PointPillars}        & L                         & \multicolumn{1}{c}{\textbf{68.61}}          & \multicolumn{1}{c}{51.26}          & \multicolumn{1}{c}{66.00}          & \textbf{62.02}          & \multicolumn{1}{c}{\textbf{90.84}}          & \multicolumn{1}{c}{\textbf{62.80}}          & \multicolumn{1}{c}{85.25}          & 79.63                   \\ 
\multicolumn{3}{c}{MLF-4DRCNet}    & R+C                       & \multicolumn{1}{c}{53.29} & \multicolumn{1}{c}{\textbf{51.36}}          & \multicolumn{1}{c}{\textbf{76.17}} & 60.28 & \multicolumn{1}{c}{90.20} & \multicolumn{1}{c}{61.83}          & \multicolumn{1}{c}{\textbf{95.66}} & \textbf{82.57}           \\ \bottomrule
\end{tabular}
\begin{tablenotes} 
\item The symbols R, C, and L denote 4D radar, camera, and LiDAR, respectively.
\end{tablenotes} 
\end{center}
\end{table*}

\subsubsection{Effect of PLFE}
We investigate the effect of PLFE by removing it and directly feeding the proposal features generated by the previous HSFP into the detection head. As shown in Table \ref{Table5}, incorporating PLFE brings AP improvements of +0.3\%, +0.71\%, and +2.7\% for the cars, pedestrians, and cyclists in the DCA, respectively. Note that the use of PLFE results in a more significant improvement for small objects, such as cyclists. We attribute this improvement to the ability of PLFE to fuse proposal-level local radar point clouds with image features. This process enhances the feature representation of small objects, which often suffer from weak feature representation in the whole scene.

\subsection{Comparison with LiDAR-based Methods}
To further demonstrate the advantages of our model, we compare it with the established PointPillars \cite{lang2019pointpillars} model in the LiDAR modality on the VoD validation set. As shown in Table \ref{Table6}, our model significantly narrows the performance gap with LiDAR-based detection in the EAA. More importantly, it outperforms the LiDAR-based PointPillars model within the DCA, which is a more critical region for driving. Specifically, our MLF-4DRCNet achieves significantly higher accuracy in detecting cyclists than PointPillars. This improvement stems primarily from radar’s unique capability to directly capture Doppler velocity information, which is highly effective for detecting moving objects such as cyclists. Moreover, we observe that within the DCA, the performance gap in car detection between our model and PointPillars is notably smaller than that in the EAA. This reduction can be explained by the increased density of radar point clouds at closer ranges. When fused with semantic features extracted from camera images, it enables our model to achieve detection performance comparable to that of LiDAR-based PointPillars.
Furthermore, given that radar and camera sensors are significantly less expensive than LiDAR sensors, these results highlight the strong cost-effectiveness and economic advantage of our proposed MLF-4DRCNet.

\begin{table}[pos=t]\rmfamily
\begin{center}
\caption{Performances of our model under different lighting conditions on the TJ4DRadSet test set.}
\label{Table7}
\setlength{\tabcolsep}{1.2mm}
\renewcommand{\arraystretch}{1.2}
\begin{tabular}{cllcccccc}
\toprule
\multicolumn{3}{c}{\multirow{2}{*}{Method}} & \multicolumn{3}{c}{3D mAP (\%)}                                                           & \multicolumn{3}{c}{BEV mAP (\%)}                                                          \\  \cmidrule(r){4-6}
\cmidrule(r){7-9}
\multicolumn{3}{c}{}                        & \multicolumn{1}{c}{Dark}           & \multicolumn{1}{c}{Normal}         & Shiny          & \multicolumn{1}{c}{Dark}           & \multicolumn{1}{c}{Normal}         & Shiny          \\ \hline
\multicolumn{3}{c}{MLF-4DRCNet-R}           & \multicolumn{1}{c}{22.43}          & \multicolumn{1}{c}{29.81}          & 26.64          & \multicolumn{1}{c}{26.13}          & \multicolumn{1}{c}{38.07}          & 33.10          \\ 
\multicolumn{3}{c}{MLF-4DRCNet}             & \multicolumn{1}{c}{\textbf{23.87}} & \multicolumn{1}{c}{\textbf{38.14}} & \textbf{32.89} & \multicolumn{1}{c}{\textbf{26.40}} & \multicolumn{1}{c}{\textbf{46.95}} & \textbf{37.16} \\ \bottomrule
\end{tabular}
\end{center}
\end{table}

\subsection{Impact of Lighting Conditions}
To investigate the impact of image quality on our model under different lighting conditions, we follow \cite{gu2025hgsfusion} in dividing the sequences of the TJ4DRadSet test set into three subsets based on scene brightness: dark, normal, and shiny, which account for approximately 15\%, 60\%, and 25\% of the test set, respectively. We then evaluate our proposed MLF-4DRCNet on these subsets. Additionally, we evaluate a variant, MLF-4DRCNet-R, which uses only raw radar point clouds as input without the HSFP and PLFE modules. As shown in Table \ref{Table7}, the fusion network outperforms the radar-only network under all lighting conditions.
It is noteworthy that in environments with significant image degradation such as dark conditions, the performance gap between MLF-4DRCNet and MLF-4DRCNet-R is relatively modest. The margin of improvement grows as image quality enhances, such as under normal lighting. This trend is expected, since poor visual conditions limit the extraction of valuable semantic cues from images for object detection. Importantly, the proposed MLF-4DRCNet maintains stable performance without exhibiting performance degradation caused by poor image quality, demonstrating its robustness to varying perceptual conditions.

\section{Conclusion}
In this study, we propose MLF-4DRCNet, a novel multi-level fusion framework that integrates 4D radar and camera at the point, scene, and proposal levels for 3D object detection. It consists of three core components: the ERPE module, which densifies and encodes radar points via the TA-VFE; the HSFP module, which performs multi-scale feature fusion with deformable attention and avoids explicit BEV representation; and the PLFE module, which refines region proposals through feature integration. Extensive experiments show that MLF-4DRCNet outperforms all existing radar-camera fusion methods on the VoD and TJ4DRadSet datasets, and achieves performance competitive with LiDAR-based models on the VoD dataset. The results demonstrate that our method offers a robust and cost-effective perception solution for autonomous driving systems operating under adverse weather conditions.

\section*{Data availability}
All the data used by this study are publicly available from the references.

\section*{Acknowledgments}
This work was supported by the National Natural Science Foundation of China under Grants 62202442 and 62471450.

% Use the style of numbering in square brackets.
% If nothing is used, default style will be taken.
%\begin{enumerate}[a)]
%\item 
%\item 
%\item 
%\end{enumerate}  

%\begin{itemize}
%\item 
%\item 
%\item 
%\end{itemize}  

%\begin{description}
%\item[]
%\item[] 
%\item[] 
%\end{description}  

% Figure
% \begin{figure}[<options>]
% 	\centering
% 		\includegraphics[<options>]{}
% 	  \caption{}\label{fig1}
% \end{figure}

% \begin{table}[<options>]
% \begin{tabular*}{\tblwidth}{@{}LL@{}}
% \toprule
% \midrule
%  & \\
%  & \\
%  & \\
%  & \\
% \bottomrule
% \end{tabular*}
% \end{table}

% Uncomment and use as the case may be
%\begin{theorem} 
%\end{theorem}

% Uncomment and use as the case may be
%\begin{lemma} 
%\end{lemma}



%\bibliographystyle{cas-model2-names}
\bibliographystyle{elsarticle-num}

\bibliography{mybib}



\end{document}